\documentclass[sigconf]{acmart}
\AtBeginDocument{%
  }

\setcopyright{acmlicensed}
\copyrightyear{2026}
\acmYear{2026}
\acmDOI{XXXXXXX.XXXXXXX}

\acmConference[xxx]{xxx}{xxx,
  xxx}{xxx}

\acmISBN{978-1-4503-XXXX-X/2026/04}

\usepackage{multirow}
\usepackage{subfigure}
\usepackage{xspace}
\usepackage{epsf}
\usepackage{setspace}
\usepackage{caption}
\usepackage{comment}
\usepackage{color}
\usepackage{algorithm}
\usepackage{algorithmicx}
\usepackage{algpseudocode} 
\usepackage{balance}
\usepackage{url,epsfig,array}
\usepackage{amsmath}
\usepackage{epstopdf}
\usepackage{cases}
\usepackage{bm}
\usepackage{color}
\usepackage{tikz}
\usepackage{threeparttable}
\usepackage{mathtools}
\usepackage{booktabs}
\usepackage{tabularx}

\usepackage{graphicx}  
\usepackage{epstopdf} % Converts EPS to PDF on-the-fly  
\def\ie{\textit{i.e.}\xspace}
\def\etal{\textit{et al.}\xspace}

\def\eg{\textit{e.g.}\xspace}

\begin{document}

\title{Improving LLM Reasoning via Dependency-Aware Query Decomposition and Logic-Parallel Content Expansion}

\author{Xianjun Gao, Jianchun Liu, Hongli Xu, Liusheng Huang}
\affiliation{
  \institution{University of Science and Technology of China}
  \country{China}
}




% \received{20 February 2007}
% \received[revised]{12 March 2009}
% \received[accepted]{5 June 2009}

\begin{abstract}
% Large Language Models (LLMs) have demonstrated remarkable reasoning capabilities, but their reasoning remains hindered by two fundamental challenges: the computational inefficiency of strictly sequential generation and rigid reasoning strategies that mismatch with query difficulty.
The integration of Large Language Models (LLMs) into real-time Web applications, such as AI-powered search and conversational agents, presents a fundamental Web infrastructure challenge: reconciling the demand for high-quality, complex reasoning with the stringent low-latency and high-throughput requirements of interactive services.
Current LLM reasoning, hindered by computationally inefficient sequential generation and rigid reasoning strategies, creates a critical bottleneck for the Web services.
Existing approaches typically optimize the LLM reasoning for either efficiency or quality but struggle to achieve both, and thus fail to meet the dual requirements of modern Web platforms.
To overcome these limitations, we propose Orion, a novel and efficient reasoning framework that enables dependency-aware query decomposition and logic-parallel content expansion. 
Concretely, Orion decomposes a single query reasoning process into two synergistic phases: (1) \textit{key point generation}, which distills logically structured key points through retrieval-augmented few-shot prompting, and (2) \textit{content parallel expansion}, which concurrently elaborates on these points based on a dependency graph to ensure logical consistency. 
Furthermore, Orion introduces a pipeline scheduling mechanism that exploits the complementary computational characteristics of the two phases (generation imposes pressure on GPU computing and expansion stresses on GPU memory) across multiple queries, enabling cross-query parallelism and dramatically improving reasoning performance (\ie, efficiency and quality).  
Experiments on diverse benchmarks show that Orion not only delivers up to 4.33$\times$ higher token generation speed and 3.42$\times$ lower answer latency over the baselines but also improves reasoning quality by up to 18.75\% through explicitly modeling inter-point dependencies. 
% These results establish Orion as a promising paradigm for achieving both efficiency and reliability in large-scale LLM reasoning.

% These challenges are particularly pressing for Web-based applications, where both efficiency and quality are essential for real-time reasoning tasks.
% However, the existing approaches typically optimize LLM reasoning for either efficiency or quality, but rarely achieve both. 
\end{abstract}

\maketitle

\section{Introduction}\label{sec:intro}
The World Wide Web is undergoing a paradigm shift, evolving from a repository of static documents into an interactive, intelligent space powered by Large Language Models (LLMs) \cite{chowdhery2023palm, rafailov2023direct,tang2025top}. 
This integration, which is redefining core Web services like search, real-time content summarization, and conversational agents, has surfaced a fundamental scientific and engineering challenge for the Web applications: the inherent tension between the Web's demand for instantaneous, high-quality information synthesis and the slow, computationally expensive nature of advanced LLM reasoning \cite{dao2022flashattention, yao2023react, plaat2024reasoning}.
At its heart, this challenge stems from the strictly sequential generation mechanism of LLMs \cite{brown2020language, minaee2024large}. 
Due to the inherent computing limitations in LLMs reasoning and high computational overhead, LLMs often take significant time to generate responses, which is unacceptable to the requirements of instantaneous web service.
In reality, many corporate departments choose to deploy LLMs on edge servers to enhance employee productivity due to data privacy and economic efficiency \cite{qu2025mobile}.
However, the high computational overhead of LLM reasoning creates a difficult trade-off for servers, such that they must decide between providing fast but potentially low-quality responses or delivering high-quality reasoning at a latency that is unacceptable for real-time Web interaction \cite{see2021understanding}.
% While this process produces coherent reasoning, it is fundamentally at odds with the performance requirements of the Web. Serving millions of concurrent users demands both extremely low latency, where delays of seconds can degrade user experience and lead to service abandonment\footnote{https://web.dev/articles/rail?hl=en}, and massive throughput. 
% The high computational overhead of current reasoning methods creates a severe bottleneck, forcing a difficult trade-off: systems can either provide fast but potentially low-quality responses, or they can deliver high-quality reasoning at a latency that is unacceptable for real-time Web interaction \cite{see2021understanding}. 
% This is not a sustainable model for the future of an AI-driven Web.
Therefore, ideal LLM reasoning requires a dual objective: \textit{efficiency}, achieved through algorithmic and system-level optimizations to minimize response time, and \textit{quality}, ensured by accurate semantic understanding and reliable contextual reasoning.

Despite significant progress \cite{bai2023qwen,team2024gemini,dubey2024llama}, current LLM reasoning optimizations remain fragmented, typically trading off efficiency against quality instead of achieving both simultaneously.
% One prominent direction is knowledge distillation \cite{xu2020bert,acharya2024survey,mcdonald2024reducing}, which transfers knowledge from large to small models to reduce computational costs. However, knowledge distillation suffers from significant information loss that degrades reasoning capability and can amplify flaws, with accuracy degradation exceeding 20\% in models like BERT or LLaMA \cite{xu2024survey}.
One prominent direction is knowledge distillation \cite{xu2020bert,acharya2024survey,mcdonald2024reducing}, which transfers knowledge from large models to smaller ones, significantly reducing the computational cost per query and decreasing reasoning time. However, distillation inherently suffers from widespread information loss, degrading reasoning capability and sometimes even amplifying flaws inherited from the large model. In models like BERT or LLaMA, the overall degradation in reasoning accuracy can be more than 20\% \cite{xu2024survey}.
Another representative line of research is multi-query parallel scheduling \cite{mei2024aios}. 
Tan \etal \cite{tan2025towards} propose optimizing overall efficiency by determining the execution priorities of different queries, thereby allowing high-priority tasks to be processed first.
In complex task scenarios, however, this method is highly susceptible to failures arising from misjudged node priorities, resulting in around 15\% decrease in overall reasoning quality.
% Furthermore, they ignored the diversity of computational characteristics in the query, causing similar ones to be scheduled onto homogeneous resources and limiting effectiveness in complex scenarios.

Several studies have aimed to improve model reasoning by enhancing generation quality for each query.
Wei \etal \cite{wei2022chain} propose Chain-of-Thought (CoT), breaking complex queries into multiple reasoning steps to guide models toward solutions and significantly improve accuracy.
Guo \etal \cite{guo2025deepsieve} propose DeepSieve, which decomposes the Retrieval-Augmented Generation (RAG) process into multiple sub-RAGs, like a directed acyclic graph (DAG), significantly enhancing reasoning quality by providing LLMs with extensive and reliable context.
Despite their demonstrated effectiveness, both methods incur significant efficiency costs.
CoT introduces significant overhead and latency due to extra calculation steps, while DeepSieve faces delays from external retrieval and semantic alignment needs (up to 3$\times$ additional time cost \cite{tan2025towards}). 
Moreover, DeepSieve neglects dependencies between RAG tasks during multi-step reasoning (\eg, travel advice should rely on the weather), which severely degrades the quality of the final reasoning answer.

% Existing technologies provide different ideas for improving the reasoning ability of LLM, but their limitations still restrict the actual application effect. 
Both aforementioned technical paths expose the defects of "Efficiency first over Quality" or "Quality first over Efficiency", and cannot fully meet the dual requirements of LLM reasoning. 
% To achieve "Efficiency at Quality" in LLM reasoning, both the inherent computational characteristics of LLM reasoning and the mismatch between the rigid reasoning strategy and problem complexity pose challenges. 
It remains challenging to achieve "Efficiency at Quality" in LLM reasoning, with hurdles stemming from two fronts.
% : the inherent computational characteristics of LLM reasoning, as well as the mismatch between the rigid reasoning strategy and problem complexity pose challenges.
\textit{First}, LLM reasoning is inherently constrained by its sequential generation. 
For a query, the model must generate tokens sequentially, with each token conditioned on its predecessors. 
This sequential dependency severely restricts parallelization, making it difficult to accelerate the reasoning process effectively.
% \textit{Second}, LLMs often suffer from the mismatch between the rigid reasoning strategy and problem complexity.
\textit{Second}, LLMs often struggle to adapt their rigid reasoning strategies to problems of varying complexity.
For example, when solving a simple arithmetic series summation, the model may laboriously add terms step by step, rather than leveraging the closed-form summation formula, which can waste a lot of computational resources. 
Conversely, for complex tasks, the model often produces incomplete or impractical results, leading to a significant decrease in output quality.
% Therefore, achieving the dual requirements of LLM reasoning still necessitates further research.
% Furthermore, in multi-query settings, it remains a challenge to design effective scheduling strategies that can exploit cross-query independence while ensuring efficient execution.
% These mismatched behaviors may lead to a dual failure of efficiency and reliability, not only wasting a lot of computing resources, but also failing to improve output quality.

\begin{figure}[t]
    \centering
    \setlength{\abovecaptionskip}{-0.2mm}
    \subfigure[Key points and their dependency]{
        \includegraphics[width=1.7in]{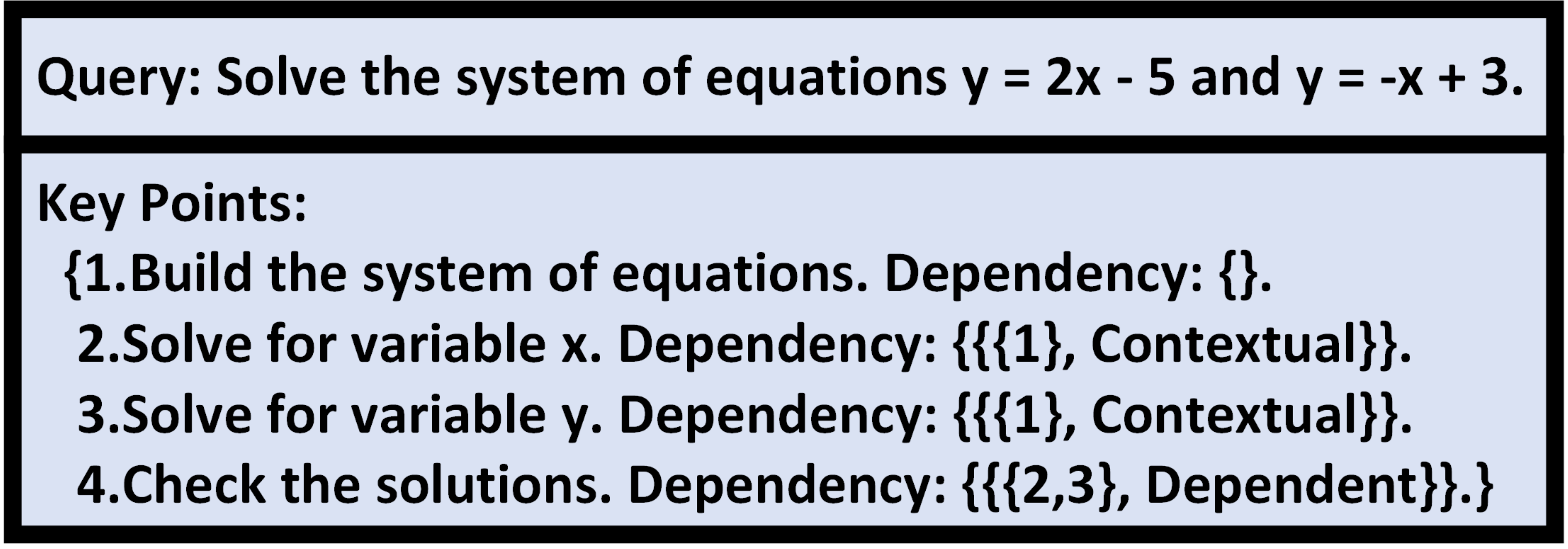}\label{subfig:relathinship}
    }\hspace{-0.1cm}
    \subfigure[Multi-query pipeline scheduling]{
        \includegraphics[width=1.5in]{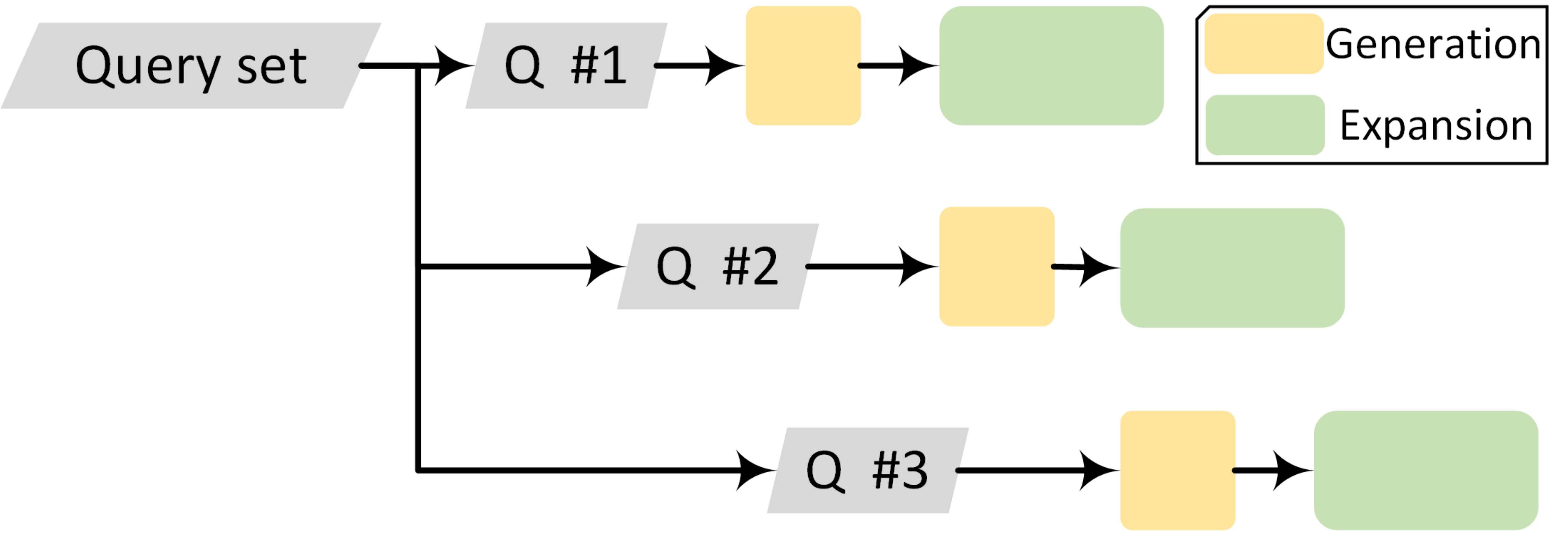}\label{subfig:pipeline_example}
    }
    \caption{(a) Orion generates four key points (\ie, build, solve x, solve y and check) for the query. Points $2$ and $3$, requiring Point $1$ as context, are independent of each other, and Point $4$ depends on the output of Points $2$ and $3$. (b) The pipeline execution between key point generation and content parallel expansion is driven by phase-specific computational traits.}
    \label{fig:test_DAG}
    \vspace{-5mm}
\end{figure}

In response to the above contradiction, we propose an efficient and novel LLM reasoning framework, named Orion, to achieve both efficiency and quality. 
The core innovation of Orion lies in its dependency-aware decomposition of complex reasoning queries in the \textit{key point generation} phase, followed by a logic-parallel expansion of the identified key points in the \textit{content parallel expansion} phase.
(1) During the key point generation phase, Orion conducts rapid semantic analysis of the input problem to extract key points and identify their interdependencies (\ie, null, contextual, and dependent). 
To maintain logical coherence while preserving parallelization benefits, we propose to implement a DAG that models inter-point dependency relationships (\eg, in Figure \ref{subfig:relathinship}, solving variables x and y requires only formulating the system of equations, whereas verifying the solution must depend on the derived values of x and y). 
(2) During the content parallel expansion phase, Orion divides each key point's content expansion into prefilling and decoding steps. 
It then constructs and optimizes a step-level DAG based on key point dependencies to enable parallel expansion.
Orion employs context-aware caching to dynamically inject relevant information from expanded points into ongoing expansions based on the DAG. This process ensures global logical consistency across all contents, ultimately guaranteeing the quality of the answers.
Furthermore, we observe that the key point generation phase is compute-intensive, placing pressure on GPU computing, whereas the content parallel expansion phase is memory-intensive, stressing on GPU memory.
Since the two phases are independent across queries and exhibit distinct characteristics, Orion enables cross-query pipeline parallelism, where subsequent queries begin key point generation without waiting for the expansion phase of preceding queries to finish, as shown in Figure \ref{subfig:pipeline_example}. 
This overlapping execution significantly reduces latency in multi-query scenarios while preserving reasoning quality.

% Based on the computational characteristics of each phase and the independence between different queries, we design a multi-query pipeline scheduling strategy and provide a brief example in Figure \ref{subfig:pipeline_example}. 
% By applying this strategy, we significantly reduce latency in multi-query scenarios without affecting the performance of individual queries.

% xxxx
% Orion not only accelerates text generation through parallelization but also ensures logical rigor via quantifying dependency relationships. 
%Speed at Quality

% If a node’s predecessor has not yet completed expansion, the node pauses execution to prevent logical inconsistencies. 
% Additionally, Orion employs context-aware caching to dynamically inject critical information from already-generated key points into ongoing expansions, ensuring that expanded content maintains local independence while globally referencing necessary contextual information.

% Throughout the parallel expansion stage, Orion enforces the order of expansion based on this DAG via a lightweight synchronization mechanism.

% This process yields structured text units with inherent dependency relationships that form the foundation for subsequent processing. 

%fully leveraging hardware parallelism to significantly reduce the overall time required for long-text generation. 
However, the implementation of Orion still faces two critical challenges. 
First, \textit{accurately identifying dependencies among key points and constructing the DAG are inherently difficult}. 
The logical relations between key points are often multi-layered and non-linear, and the model’s identification of such dependencies relies heavily on its intrinsic reasoning ability.
Misclassification of these relationships can propagate errors into the expansion phase, leading to degraded quality or even entirely flawed reasoning.
Second, the two phases of Orion exhibit markedly different computational characteristics. 
The key point generation phase is compute-intensive, requiring deep contextual analysis and sequential reasoning, whereas the content parallel expansion phase is memory-intensive, demanding frequent access to GPU memory for expanding multiple key point contents.
This disparity can lead to severe resource underutilization, where one resource (\eg, GPU computing or GPU memory) becomes a bottleneck while the other sits idle.
Thus, \textit{designing an effective scheduling strategy that leverages these heterogeneous characteristics across different queries remains a key challenge} for enhancing execution efficiency in multi-query settings.

The main contributions are summarized as follows:
\begin{itemize}
    \item We propose Orion, a novel LLM reasoning framework that breaks down the reasoning process for a complex query into two synergistic phases: key point generation and content parallel expansion. By leveraging the distinct computational characteristics of each phase, Orion enables cross-query phase parallelism, achieving a principled balance between efficiency and quality.
    \item We design a dependency-aware parallel expansion algorithm modeling inter-point relationships with a DAG and integrate context-aware caching to ensure logical coherence while enabling efficient parallel content expansion.
    \item Experiments on diverse benchmarks show that Orion not only delivers up to 4.33$\times$ higher token generation speed and 3.42$\times$ lower answer latency over the baselines, but also improves reasoning quality by up to 18.75\% through explicitly modeling inter-point dependencies.
\end{itemize}

\section{Motivations for System Design}\label{sec:prelim}

\begin{figure}[t]
    \centering
    \setlength{\abovecaptionskip}{-0.2mm}
    \subfigure[Token generation speed-up ratio]{
        \includegraphics[width=1.75in]{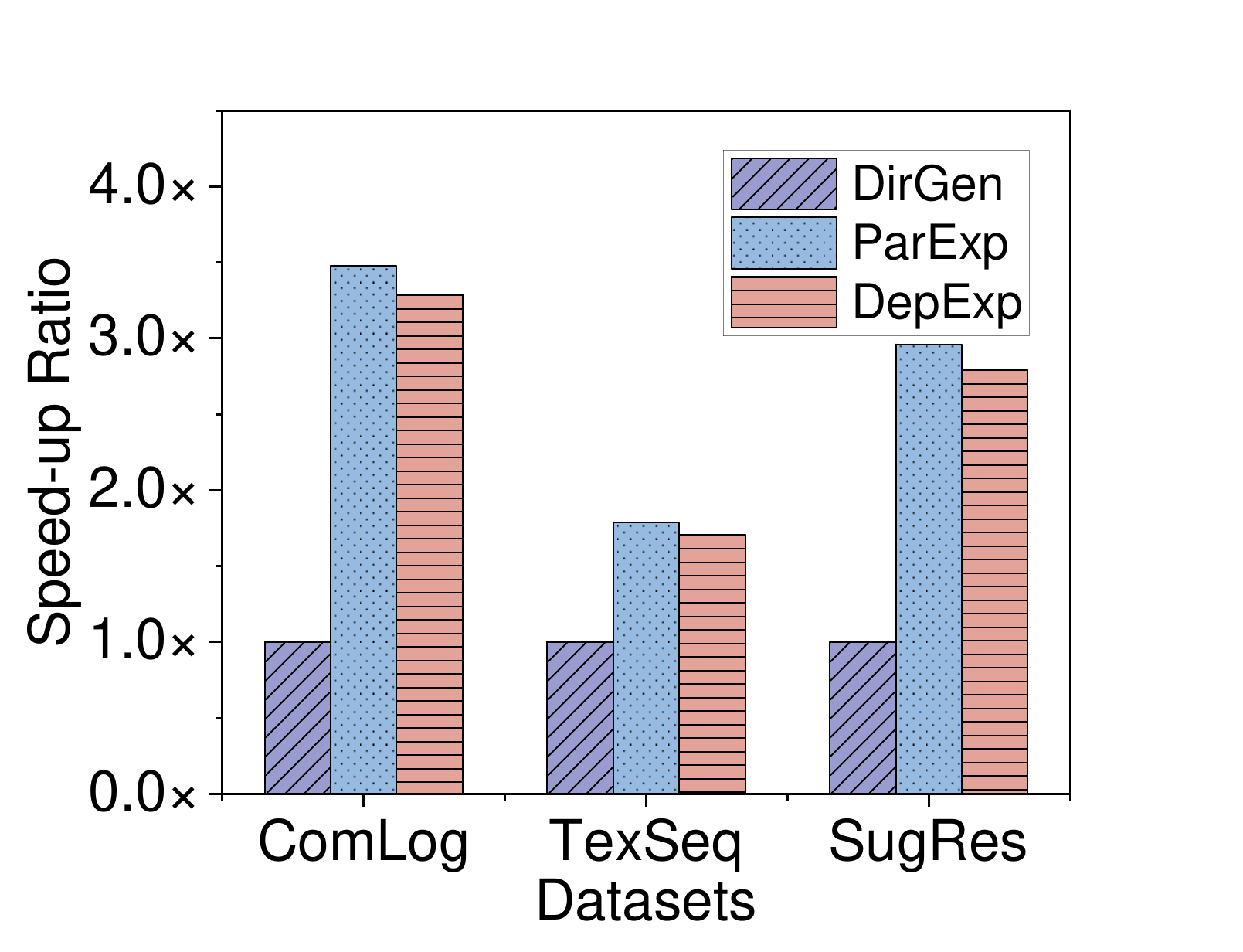}\label{subfig:test_speed_up_ratio}
    }\hspace{-0.9cm}
    \subfigure[Win rate against DirGen]{
        \includegraphics[width=1.75in]{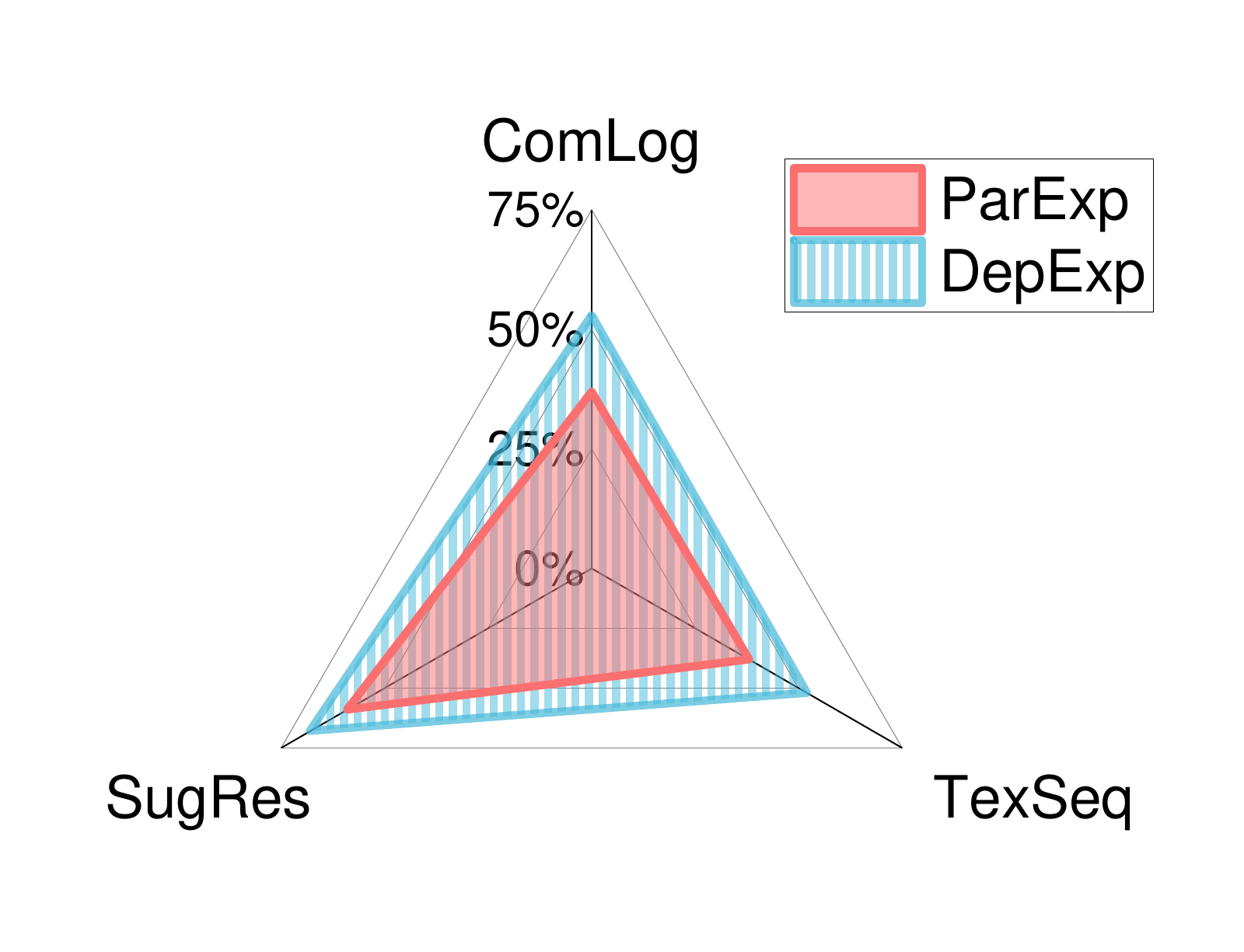}\label{subfig:test_text_qua}
    }
    \caption{The reasoning efficiency and quality of different methods on three datasets.}
    \label{fig:test_independent}
    \vspace{-8mm}
\end{figure}

LLM reasoning typically adopts a sequential generation strategy, producing output sequences token by token, with each new prediction strictly conditioned on all previously generated tokens. 
This strong sequential generation introduces substantial computational latency, as subsequent tokens cannot be computed until their predecessors are finalized, thereby limiting the hardware potential for parallel computation.
In contrast, humans often address complex questions by first outlining key points and an overarching logical structure before elaborating on each point independently.
Inspired by this process, we propose enabling LLMs to move beyond rigid sequential generation constraints by generating multiple \textit{logically independent} and \textit{semantically parallel} content segments concurrently. 
When such segments exhibit weak or no interdependence, parallel generation can preserve global coherence and quality while significantly enhancing reasoning efficiency.

\subsection{Impact of Inter-point Dependency}
\textbf{Different strategies for generating and sequencing key points lead to varying impacts on both coherence and correctness of the expanded answers.} 
To evaluate the feasibility of generating key points and then expanding the content of each point in parallel, we design a set of controlled experiments.
Considering that logical relationships among key points may vary across queries, we construct multiple datasets to examine how different dependency patterns affect generation outcomes.
Specifically, we manually partition the Vicuna and WizardLM datasets \cite{chiang2023vicuna,xu2023wizardlm} into three subsets: SugRes, TexSeq, and ComLog.
1) In the SugRes dataset \cite{ning2023skeleton}, key points in generated answers are mutually independent. 
2) For the TexSeq dataset \cite{ning2023skeleton}, the expansion of point content requires contextual correspondence between preceding and subsequent content. 
3) In the ComLog dataset \cite{ning2023skeleton}, key points exhibit strict dependencies where correct calculations rely on preliminary computational results, making parallel processing of all points prone to errors.  
We compare two distinct text generation strategies with the basic method and the enhanced method, respectively.
The basic method adopts a two-phase pipeline: first generating key points, then independently expanding each in parallel. 
In contrast, the enhanced method integrates an innovative logical correlation analysis mechanism, which actively identifies and exploits potential interconnections among key points after their initial generation. 
For example, in event planning, the basic method expands “determining the event theme” and “designing the opening segment” independently, while the enhanced method links thematic info to the opening expansion for consistency.
This makes the enhanced method produces more context-adaptive outputs, without losing parallel efficiency.

For clarity, we refer to the approach utilizing only parallel expansion of the point content as \textit{ParExp}, and the approach that considers the relationships between key points during expansion as \textit{DepExp}. 
We establish \textit{DirGen}, a direct sequential generation method, as our experimental baseline. 
In this setup, Llama2-7B \cite{touvron2023llama} processes input queries and generates answers without any additional processing.
As shown in Figure \ref{subfig:test_speed_up_ratio}, compared to DirGen, both ParExp and DepExp can significantly accelerate text generation speed. 
For instance, on the ComLog dataset, ParExp and DepExp achieve token generation speed 3.48$\times$ and 3.29$\times$ higher than that of DirGen, respectively, greatly enhancing output efficiency.

Furthermore, we observe that DepExp is slightly slower than ParExp in generation speed. 
This is because DepExp introduces additional computational overhead to manage inter-point dependencies during text expansion. Nevertheless, the benefits of this extra processing are evident. 
% As shown in Figure \ref{subfig:test_text_qua}, DepExp consistently outperforms ParExp in text quality across all datasets.
As shown in Figure \ref{subfig:test_text_qua}, we evaluate the win rate of ParExp and DepExp against DirGen, and found that DepExp consistently outperforms ParExp in text quality across all datasets.
Notably, on the ComLog and TexSeq datasets, ParExp yields weaker quality than DirGen, whereas DepExp surpasses DirGen, highlighting the importance of modeling dependency relationships in maintaining logical coherence. By contrast, on the SugRes dataset, DepExp offers several improvements over ParExp, since the key points are inherently independent, allowing parallel expansion without quality loss.
These findings motivate us to accurately identify and leverage dependency relationships between key points, enabling the maintenance of logical coherence and global consistency during parallel content expansion while ultimately accelerating text generation speed and enhancing output quality.

\subsection{Opportunity for Cross-query Scheduling}
\begin{figure}[t]
    \centering
    \setlength{\abovecaptionskip}{-0.2mm}
    \subfigure[Two-phase parallel scheduling]{
        \includegraphics[width=1.5in]{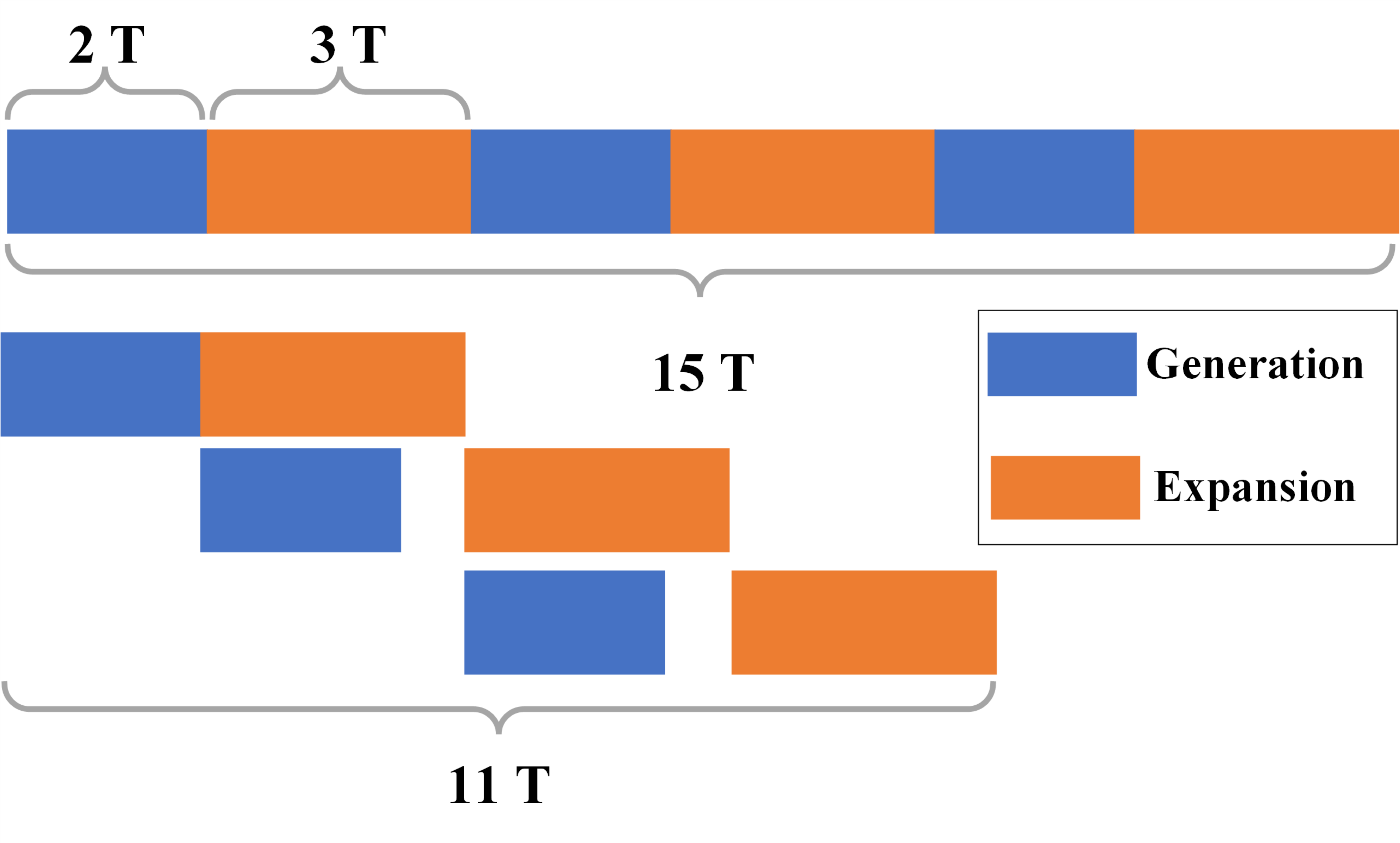}\label{subfig:relationship_example}
    }\hspace{0.1cm}
    \subfigure[Speed-up ratio and win rate]{
        \includegraphics[width=1.6in]{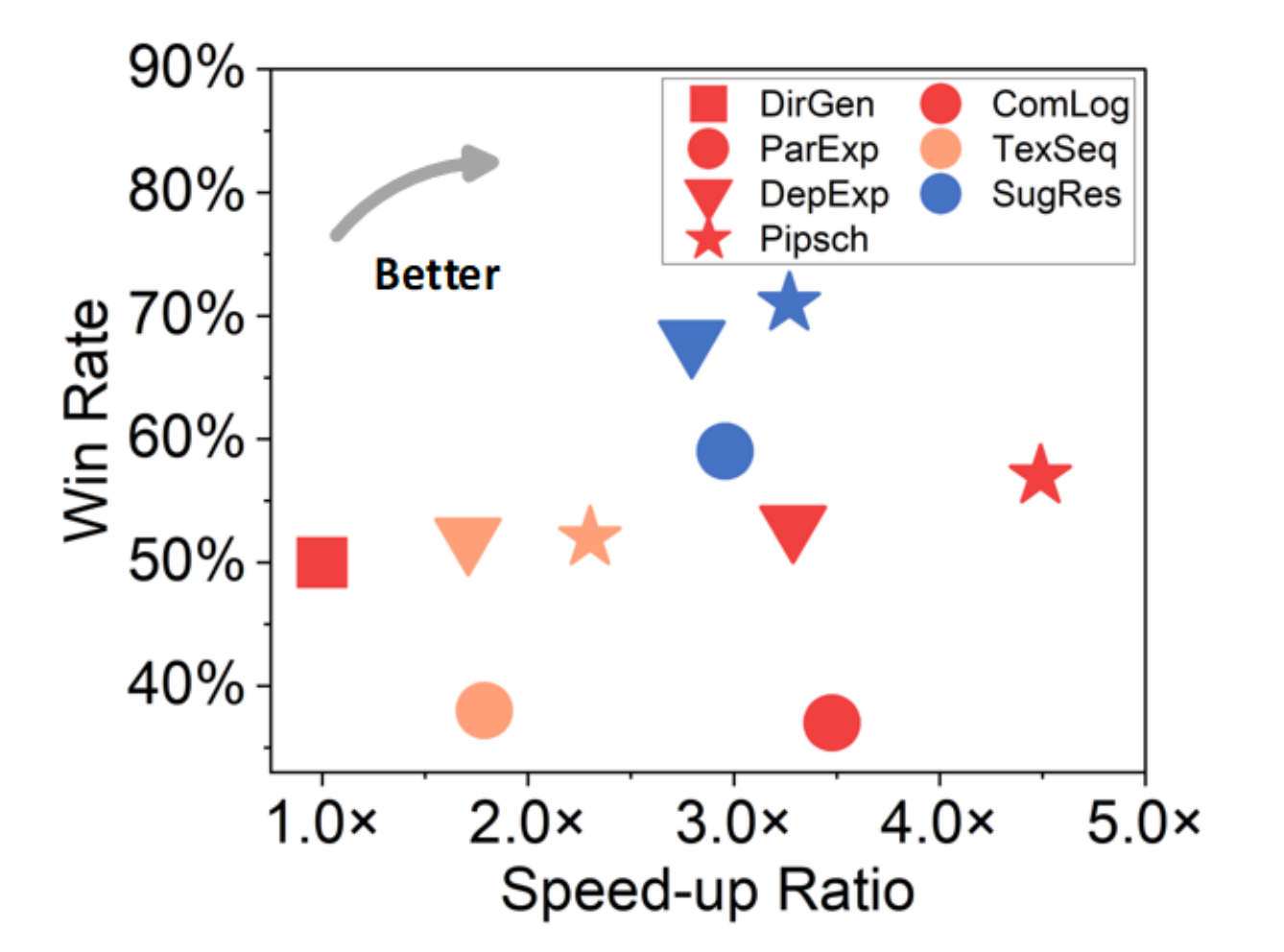}\label{subfig:Speed_up_Win_rate_Qua}
    }
    \caption{(a) Generation (stressing GPU computing) and extension (stressing GPU memory) phases for different queries can execute in parallel. (b) Shapes denote methods and colors represent reasoning performance across datasets.}
    \label{fig:test_DAG}
    \vspace{-5mm}
\end{figure}

\textbf{Leveraging the cross-query independence and computational characteristics of two phases can reduce query completion time under multiple queries.}
We further analyze the requirements of the generation and expansion phases. 
During key point generation, the model must process substantial contextual information to distill concise and relevant key points. This results in a pattern of long input and short output, making this phase computationally intensive.
Conversely, the parallel expansion phase treats each key point as an independent query, characterized by short inputs and long outputs. To achieve content expansion, it requires frequent access to the KV cache \cite{hooper2024kvquant, liu2023scissorhands} stored in GPU memory, thus exhibiting memory-intensive behavior.
Importantly, the key point generation of one query and the parallel expansion of another are independent, introducing opportunities for concurrent execution. 
This complementarity between two phases motivates us to explore whether jointly scheduling them can improve task parallelism or not in multi-query scenarios. 
% Specifically, we integrate the content parallel expansion of the previous query with the key point generation phase of the subsequent query and provide them together to the model. Since the key point generation phase primarily stresses GPU computation, while content parallel expansion mainly pressures GPU memory, performing only content parallel expansion leaves the GPU computational capacity relatively idle. This two-phase parallel approach fully utilizes different resources on the GPU.
% We illustrate the system scheduling design in Figure \ref{subfig:relationship_example}, and refer to this multi-query strategy as \textit{Pipsch}.
Specifically, we integrate the content parallel expansion of a previous query with the key point generation of a subsequent query, providing both to the model concurrently. As illustrated in Figure \ref{subfig:relationship_example}, we term this multi-query scheduling method as \textit{Pipsch}. This design efficiently utilizes different GPU resources since key point generation primarily stresses GPU computation, while content parallel expansion mainly pressures GPU memory. By executing these phases in parallel, the method exploits both the computational and memory capacities of the GPU and mitigates resource idle time.

% Regarding the generation and expansion of key points for questions, the completeness of the generated points and the appropriateness of their dependencies significantly impact the quality of the final output. However, in DepExp, we solely rely on the LLM itself to understand the question and generate the corresponding key points for the answer. Can this method, which relies solely on the LLM itself to generate key points and their dependency relationships, truly guarantee the completeness of the points and the correctness of their relationships? To investigate this question, we examined the key points generated by DepExp for the question. As illustrated in Fig. \ref{subfig:relationship_example}, taking the example problem of solving a system of equations shown, the key points generated solely by the LLM lack a step for calculating X. Furthermore, the LLM did not recognize that variables X and Y could be solved simultaneously. This resulted in an output that is not only missing part of the solution but also reduced the parallelizability of the generated key points, lowering generation efficiency. Based on this finding, we manually supplemented the missing key points for each problem and established correct dependency relationships for these supplemented points. We then re-ran the experiment and refer to this method as RelVar.

As illustrated in Figure \ref{subfig:Speed_up_Win_rate_Qua}, system efficiency is substantially improved when the expansion of point content from the previous query is executed concurrently with the generation of key points for the next. For instance, on the ComLog dataset, Pipsch (\ie, red pentagram) achieves a 4.49$\times$ speedup, markedly outperforming the 3.48$\times$ and 3.29$\times$ speedups of ParExp (\ie, red circle) and DepExp (\ie, red inverted triangle). Moreover, Pipsch maintains answer quality on par with DepExp, and in some cases even yields slight improvements. These findings strongly validate our design hypothesis. 
Based on this, we establish the overall framework of the proposed system: for each single query, the reasoning process is decomposed into two phases, \ie, key point generation and content parallel expansion. 
Across multiple queries, the proposed system leverages the distinct computational characteristics and intrinsic independence of these phases to adopt a cross-query parallel scheduling strategy, thereby achieving both enhanced efficiency and improved quality in multi-query environments.

\section{System Design of Orion}\label{sec:system}
% \begin{figure*}[t]
%     \centering
%     \includegraphics[width=1\textwidth]{fig/System/Version1.pdf}
%     \caption{System}
%     \label{System_framework}
% \end{figure*}

% \begin{figure*}[h]
%     \centering
%     \includegraphics[width=1\textwidth]{fig/System/DAG+explosion.pdf}
%     \caption{The workflow of key point generation and content parallel expansion for a single query}
%     \label{Content_Parallel_Expansion}
%     \vspace{-3mm}
% \end{figure*}

\subsection{System Overview}\label{subsec:Architecture} 
In this section, we introduce the Orion framework, which is designed to apply a query decomposition strategy to partition complex reasoning queries into two complementary phases, \ie, key point generation and content parallel expansion.
In the key point generation phase, Orion integrates related-query retrieval with a few-shot learning paradigm, ensuring that the extracted key points are both logically coherent and tightly aligned with the query statement.
% This provides a solid foundation for efficient downstream processing. 
In the subsequent content parallel expansion phase, Orion enables concurrent elaboration of expandable key point content while maintaining logical consistency across each point, thereby substantially accelerating the solution generation. All expanded content results are ultimately merged sequentially into a complete answer.
Through this two-phase design, Orion successfully achieves efficient parallel processing without sacrificing the logical coherence of the output, effectively mitigating the inherent inefficiencies of traditional sequential reasoning.

Moreover, for multi-query scenarios, Orion incorporates advanced pipeline optimization techniques to significantly boost overall throughput and computational efficiency. 
Since the key point generation and content parallel expansion phases are functionally and computationally independent across different queries, Orion enables cross-query pipeline parallelism. 
Specifically, the key point generation of a subsequent query can begin immediately after the same phase of the preceding query completes, without waiting for the expansion phase of the previous query to be completed. 
This overlapping execution forms a continuous processing pipeline in which different queries progress concurrently through different phases. 
By keeping computational resources fully engaged, the pipeline design dramatically reduces inter-query waiting time and maximizes system utilization. 
As a result, Orion achieves remarkable efficiency in managing diverse concurrent queries while maintaining logically coherent and contextually consistent answers.

\subsection{Key Point Generation}\label{subsec:key_point} 
In the key point generation phase for each query, Orion aims to transform the original query into a set of logically interconnected key points through a single reasoning step. 
To ensure the quality and logical consistency of generated key points, Orion uses a retrieval tool to search for relevant content from the database \cite{gao2023retrieval, guan2025deeprag}.
While more retrieved content can improve reliability, excessive or unrefined results may introduce errors and reduce output quality.
Additionally, processing large amounts of retrieved data during LLM prefilling increases computational load and slows generation, making it unsuitable for latency-sensitive scenarios.
To balance retrieval volume and quality, the system converts the retrieved text into a vector database for further processing.
Orion first segments the obtained text content into smaller chunks suitable for language model processing, then converts the text into embeddings and supplements them into the vector database \cite{yi2023simple, wu2024image}.
This transforms all vectors into a searchable index structure. 
Then, Orion automatically establishes a mapping relationship between vectors and the original text, returning a query-ready vector database object.
By calculating similarity \cite{caspari2024beyond, liu2023finch, arslan2024survey}, Orion obtains the most relevant text to the query content and ultimately returns these chunks along with their corresponding original text information.

% In order to ensure the quality and logical correctness of the generated key points, Orion first calls the retrieval tool to search for relevant content in the database after receiving a problem. 
% While a larger volume of retrieved content may enhance the reliability and openness of key point, excessive or unrefined results risk introducing erroneous information that could compromise output quality. 

After obtaining search results, the system integrates the retrieved content with the original query and predefined prompt to form a comprehensive input for the LLM prefilling. Within this process, the system provides diverse exemplars to guide the model to learn and comprehend both the required content and the expected generation format. This method not only improves the correlation between generated key points and queries, but also ensures that the output of the model is closely consistent with the requirements for further processing.
During point generation, the system continuously produces tokenized output units. Upon generating coherent outputs, these are immediately relayed to the next phase, mitigating delays caused by waiting for full decoding completion. Throughout this process, the LLM reasoning must structure its output in a segmentable format (\eg, JSON), explicitly indicating that the content can be semantically divided into distinct points. A dedicated parser monitors the decoding stream for structured output fragments in real-time; once a complete decodable segment is detected, it is promptly extracted and forwarded to downstream processes.

\begin{algorithm}[t]
    \caption{DAG-guided content parallel expansion} \label{alg_DAG}
    \begin{tabular}{@{}p{\columnwidth}@{}}
        \textbf{Input:} The generated key point content $P_i$ and the relationship set $Rel(i)$ for each key point $i$. \\
        \textbf{Output:} The answer for the current query.
    \end{tabular}
    \begin{algorithmic}[1]
        \State Initialize DAG $G = \{V, E\}$ for node set $V$ and edge set $E$. {\color[RGB]{148,0,211}\algorithmiccomment{Construct and optimize DAG}} \label{alg1_1}
        \For{each node $i$} \label{alg1_2}
            \State Add node $i$ in the node set $V$ and assign the key point content $P_i$ for node $i$. \label{alg1_3}
        \EndFor \label{alg1_4}
        \For {each dependency point $j \in Rel(i)$} \label{alg1_5}
            \State Add the directed edge $i \xrightarrow{} j$ in the edge set and assign the relationship type for the edge. \label{alg1_6}
        \EndFor \label{alg1_7}
        \State Split each node into prefilling and decoding node and optimize the DAG. \label{alg1_8}
        \State Initialize the running node set $R$. {\color[RGB]{148,0,211}\algorithmiccomment{Content parallel expansion}} \label{alg1_9}
        \For{each node $i \in V$} \label{alg1_10}
            \If {node $i$ has no incoming edges} \label{alg1_11}
                \State Extract the key point content $P_i$ and combine it with the preset prompt as Eq. \eqref{input} to form the complete input. \label{alg1_12}
                \State Add the node $i$ in the running node set $R$. \label{alg1_13}
            \EndIf \label{alg1_14}
        \EndFor \label{alg1_15}
        \State LLM processes the running node set $R$. \label{alg1_16}
        \While {there is unexecuted node $j$} \label{alg1_17}
            \If {all the parent nodes for node $j$ finish} \label{alg1_18}
                \State Extract the key point content $P_j$ and combine it with all the parent nodes and the preset prompt as Eq. \eqref{input} to form the complete input. \label{alg1_19}
                \State Add the node set $j$ to the running node set $R$.  \label{alg1_20}
            \EndIf \label{alg1_21}
        \EndWhile \label{alg1_22}
    \end{algorithmic}
\end{algorithm}

\begin{figure*}[h]
    \centering
    \includegraphics[width=1\textwidth]{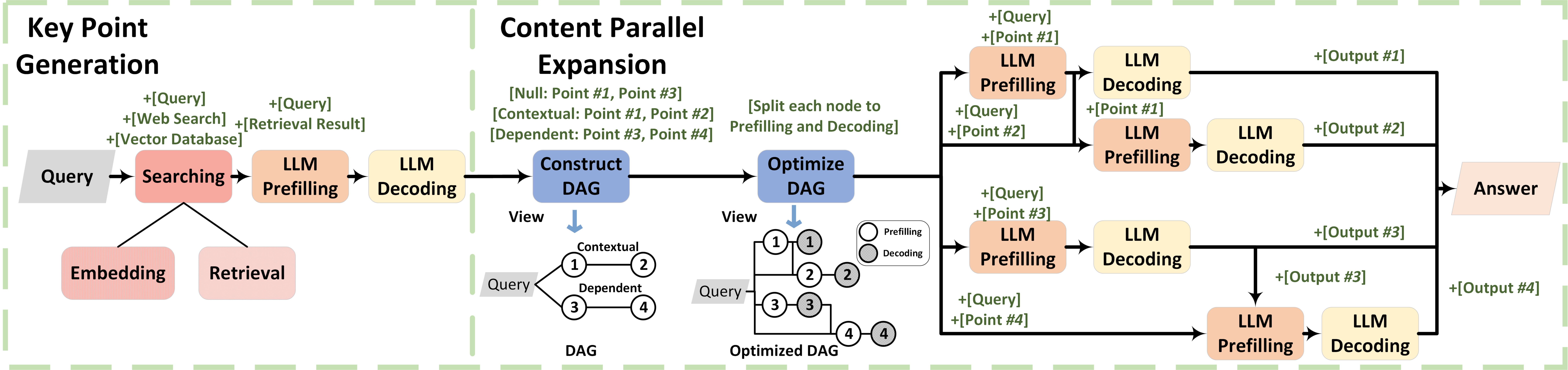}
    \caption{The workflow of key point generation and content parallel expansion for a single query}
    \label{Content_Parallel_Expansion}
    \vspace{-3mm}
\end{figure*}
\subsection{Content Parallel Expansion}\label{subsec:content_parallel}
After generating all the key points and their dependencies, Orion enters the content parallel expansion phase. 
To facilitate understanding, we propose the DAG-guided content parallel expansion algorithm, as shown in Alg. \ref{alg_DAG}. 
This algorithm first transforms the generated key points and their dependencies into a DAG that describes the points and their logical relationships. Then, based on this DAG, Orion achieves content parallel expansion of each point. 

At first, Orion initializes a graph structure where each node in the graph represents a key point and is assigned corresponding point content (Line \ref{alg1_1}). Directed edges are then established between nodes based on the provided relationship set $Rel(i)$ (\eg, from node $i$ to its dependent node $j$) (Lines \ref{alg1_2}-\ref{alg1_7}), thereby constructing a DAG that reflects the dependency relationships between key points. In this DAG, there are three distinct types of relationships between nodes: Null, Contextual, and Dependent. An None relationship indicates that no direct logical connection exists between the two key points. A Contextual relationship means that the content of the subsequent node needs to correspond to the preceding node. Here, the subsequent key point only requires awareness of the preceding key point's information and does not need to wait for its complete output. When the relationship is Dependent, the subsequent node must wait for the full output from the preceding node and can only generate its result based on that complete answer. After constructing the DAG, the system needs to further split each node into preﬁlling and decoding nodes, and then optimize the DAG based on the types of edge relationships (Line \ref{alg1_8}).
By defining these relationships between different key points, it provides guidance for the subsequent content parallel expansion.

The content expansion of the algorithm starts by identifying all nodes in the graph with zero in-degree (\ie, nodes without any predecessor dependencies) as the following formula:
\begin{equation} \label{in_degree}
	R_0 = \{ i \in V \mid \operatorname{deg}^{-}(i) = 0 \}
\end{equation}
where $\operatorname{deg}^{-}(i) = |\{j \mid (j,i) \in E \}|$ represent the in degree of node $i$. Orion sets up a running set $R$ to store all execution nodes, into which all nodes with zero in-degree should be placed for content expansion (lines \ref{alg1_9}-\ref{alg1_15}).
Since the nodes in the running set $R$ have no predecessor dependencies at this point, the corresponding key points are logically independent of each other in the answer's logic, meaning these key points can be executed in parallel. 
For all nodes in the running set $R$, their key point content is combined with a preset prompt to form a complete input as follows:
\begin{equation} \label{input}
\operatorname{Input}_j = \operatorname{Concat}\left( 
    \operatorname{Prompt}_{\text{Pre}}, 
    \textstyle\bigoplus_{k \in \operatorname{Par}(j)} f(k,j), 
    P_j 
\right)
\end{equation}
\vspace{-5mm}
\begin{equation} \label{Agg}
f(k,j) = 
\begin{cases} 
    P_k , & \operatorname{E}(k,j) = \texttt{Contextual} \\ 
    \operatorname{Output}_k , &  \operatorname{E}(k,j) = \texttt{Dependent}
\end{cases}
\end{equation}
where $\operatorname{Par}(j)$ is the parent nodes for node $P_j$ and $\operatorname{Prompt}_{\text{Pre}}$ is the preset prompt. We use $\operatorname{Concat}(\cdot)$ to denote the splicing operation. Furthermore, $\textstyle\bigoplus$ denotes the aggregation operation, and $f(k,j)$ represents the information provided by point $k$. If the relationship between $k$ and $j$ is contextual, only the point content $P_k$ need to be provided, whereas if it is a dependency relationship, the complete output result $\operatorname{Output}_k$ is required.
This whole process adopts the concept of topological sorting, initiating the entire workflow by first processing the independent nodes. Subsequently, the algorithm enters a loop, continuously checking for any unexecuted nodes. A node is added to the running set $R$ only when all the key points it depends on have been executed (Lines \ref{alg1_17}-\ref{alg1_22}), thereby ensuring the logical correctness of task execution and adherence to dependency constraints. This process will continue until all nodes have been executed.

In order to illustrate the content parallel expansion phase, we take Figure \ref{Content_Parallel_Expansion} as an example. Assume there are four key points generated for the current query in the Key Point Generation phase. Among them, Key Point $\#1$ and $\#3$ are independent, $\#1$ and $\#2$ have a contextual relationship, and $\#4$ depends on the output of $\#3$. First, a DAG is built and optimized based on Alg. \ref{alg_DAG}, as shown on the left side of the figure, to visualize these relationships. During parallel expansion execution, the system simultaneously initiates the LLM Prefilling step (orange part) for $\#1$ and $\#3$. Upon completion, given the contextual relationship between $\#1$ and $\#2$, the LLM Prefilling for $\#2$ immediately incorporates the information of $\#1$ as additional input and executes in parallel with the LLM Decoding step (yellow part) for $\#1$ and $\#3$. Once the Prefilling for $\#2$ is completed, system proceeds to the LLM Decoding step for the point content expansion. For $\#4$, the system waits for the LLM Decoding of $\#3$ to finish, then incorporates the content of $\#3$ as input into the LLM Prefilling for $\#4$, followed by its Decoding step. All outputs are finally merged into a complete answer.
%The entire workflow clearly illustrates the paths of parallel and sequential execution through color coding and arrows, ensuring efficient processing.

\subsection{Multi-query Scheduling Optimization}\label{subsec:multi_task}
After determining the complete execution process of a single query, we further optimize multi-query scheduling based on the distinct computational characteristics of two phases in each query. At the beginning of the key point generation phase, the searching operation often transforms and searches via a vector database. This process involves extensive vector searches which occupies memory. During the actual key point generation, the LLM's output is typically short, but it requires processing very long contextual information to understand the task requirements. Here, the LLM's input is much longer than its output, making this phase a compute-intensive task that occupies GPU computing units. In the content parallel expansion phase, the content expansion of each key point requires frequent access to the KV Cache stored in the GPU memory, making this phase a memory-intensive process that exerts pressure on the GPU memory.

Based on the different computational characteristics of each phase, we parallelize programs with heterogeneous computational properties across different queries to increase the overall task concurrency of the system. For instance, once the searching of a previous query is completed, the LLM prefill of that query and the searching of the next query can proceed simultaneously. Furthermore, when the searching, LLM prefill, and LLM decoding of three different queries are executed in parallel, their respective hardware pressures are on memory, GPU computing units, and GPU memory bandwidth. In this way, by considering the computational characteristics and hardware pressures of each subprogram within different queries, we can assign various subprograms from different queries to different hardware components. This approach not only prevents hardware idling but also enhances multi-query parallelism and reduces overall task latency.

\begin{figure}[t]
    \centering
    \includegraphics[width=0.5\textwidth]{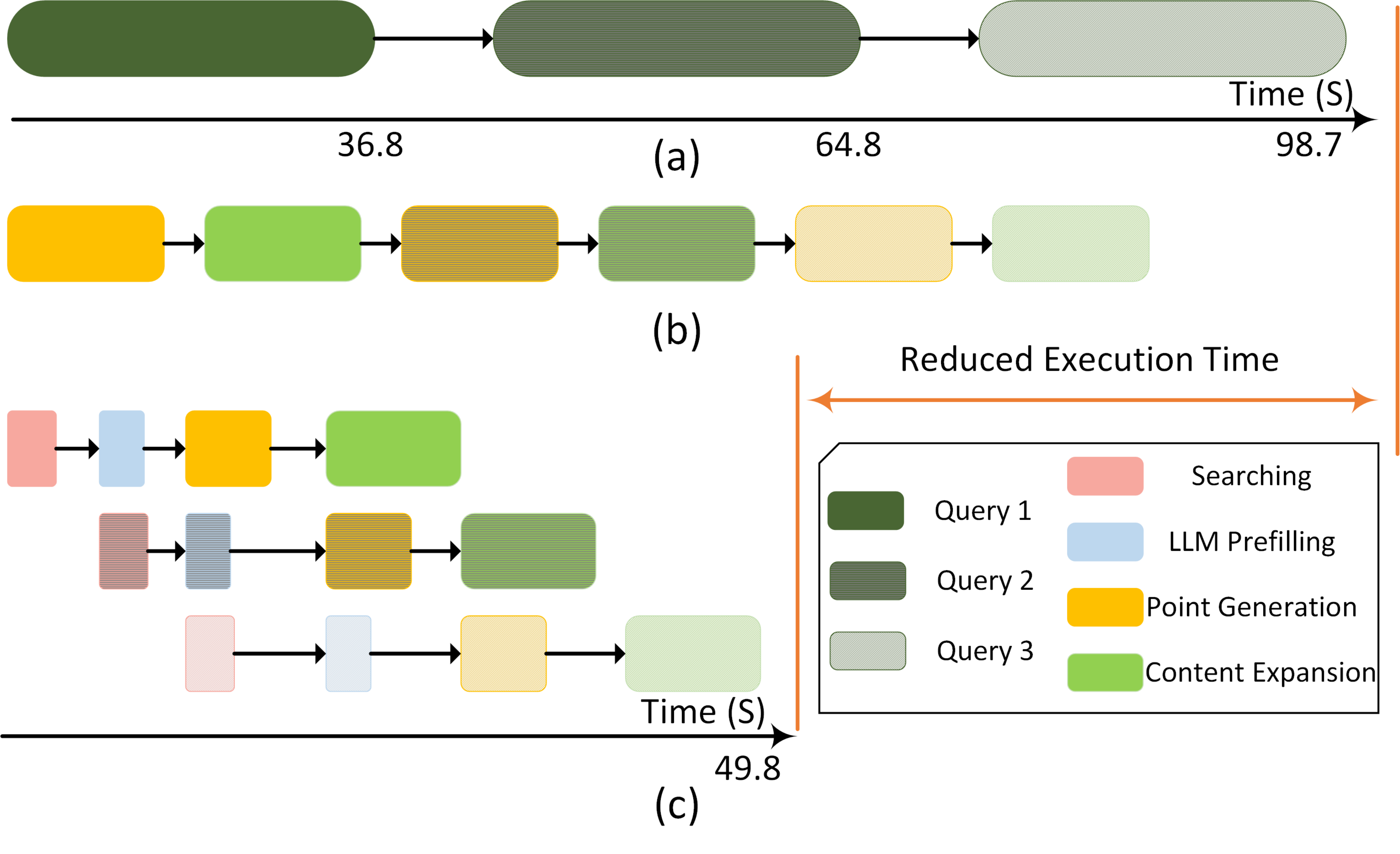}
    \vspace{-8mm}
    \caption{Workflow expression and execution comparison of existing reasoning strategy and Orion. (a) Query-level workflow in current reasoning strategy. (b) Decomposition-based dataflow graph in Orion (Searching and Prefilling are omitted due to space constraints). (c) Execution graph after multi-query scheduling optimization in Orion.}
    \label{Gantt}
    \vspace{-7mm}
\end{figure}

% \begin{figure*}[t]
%     \centering
%     \includegraphics[width=1\textwidth]{fig/System/Version1.pdf}
%     \caption{System}
%     \label{System_framework}
%     \vspace{-3mm}
% \end{figure*}

We explain and compare the workflow between the existing schemes and our framework in Figure \ref{Gantt}. The existing reasoning strategy \cite{wu2023autogen,hao2024training} naturally adopts query-level reasoning optimization method (Figure \ref{Gantt}(a)), where each query is viewed as an independent unit and ignores potential optimization possibilities that query can be decomposed. We excavate the decomposability of queries and explore the logical relationships between various parts after decomposition. By first breaking down the query into multiple key points and then simultaneously processing the logically parallelizable parts at the same moment based on the logical relationships between different points (Figure \ref{Gantt}(b)), we accelerate the reasoning efficiency for each individual query. Furthermore, we investigate the computational characteristics of each phase during query execution and discover that the characteristics of different phases are inconsistent. This allows us to more effectively leverage the correspondence between different computational characteristics and the system hardware to achieve cross-query phase parallel acceleration in the multi-query scenario (Figure \ref{Gantt}(c)). Through this strategy, we can ultimately realize parallel reasoning acceleration in multi-query settings based on the task execution characteristics of different phases. 

\section{Experimental Evaluation}\label{sec:evaluation}
\subsection{Datasets and Models}\label{subsec:dataset_model}

\textbf{Datasets}: We evaluate and compare the reasoning performance of existing methods and Orion using two recent assistant-style datasets:
(1) Vicuna \cite{chiang2023vicuna}, which contains a total of 80 queries covering 9 diverse categories, including advice, mathematics, writing, coding and so on.
(2) WizardLM \cite{xu2023wizardlm}, which includes 218 queries spanning more categories (\eg, advice, writing and coding) and varying difficulty levels (ranging from simple speed calculations to solving complex differential equations).

\textbf{Models}: We test Orion and existing methods on three different open-source models: LLaMA2 7B \cite{touvron2023llama}, LLaMA2 13B \cite{touvron2023llama}, and Qwen2.5 7B \cite{qwen2.5}.
We obtain the weights of all open-source models from Hugging Face\footnote{https://huggingface.co/}. 
These three models are selected to compare Orion's performance across the same architecture but with different parameter scales (\ie, LLaMA2 7B and LLaMA2 13B) and across different model architectures with similar parameter scales (\ie, LLaMA2 7B and Qwen2.5 7B).

\subsection{Evaluation Setup}\label{subsec:experiment_setup}
\textbf{Baselines:} We adopt three different baselines for performance comparison: (1) \emph{Normal} generates reasoning results sequentially by feeding the original query directly into the model for reasoning. It produces tokens sequentially, with each new token conditioned on all previously generated tokens until an end token is emitted or the maximum length is reached. 
(2) \emph{CoT} \cite{wei2022chain} guides LLM to demonstrate human-like, step-by-step reasoning through carefully designed prompts, thereby improving the model's reasoning quality. 
This approach helps the model break down complex problems into multiple intermediate computational steps, which are then processed sequentially to arrive at the final answer. 
(3) \emph{SoT} \cite{ning2023skeleton} guides LLMs to first quickly generate the skeleton of an answer and then fill in the detailed content within this skeleton, significantly enhancing the efficiency of model reasoning. 
However, this method only avoids compromising answer quality when dealing with problems where the skeleton points are highly independent of each other.

\textbf{Performance Metrics:} We mainly employ the following metrics to evaluate the performance of different methods: (1) \textit{Token Generation Speed-up Ratio}. In each query, we record the number of tokens generated by the model and the required time, and calculate the speed-up ratio based on the number of generated tokens per second achieved by each method relative to Normal. (2) \textit{Win Rate}. To compare the quality of answers generated by different methods, we employ an evaluation framework based on two distinct judge models (\ie, GPT-5 \cite{achiam2023gpt} and DeepSeek \cite{liu2024deepseek}). 
For each comparison, the evaluation framework presents the model with a set of queries and two corresponding answers for each query, soliciting its preference between the two answers for every query. (3) \textit{Latency}. We measure the time from initiation to the completion of the entire reasoning process for each dataset, in order to evaluate and compare the total time required by different methods to answer all queries.

\textbf{Platform:} Our experiments are conducted on a deep learning workstation, which is equipped with an Intel(R) Xeon(R) Platinum 8358P, 4 NVIDIA RTX A6000 GPUs and 512 GB RAM.

\begin{figure}[t]
    \centering
    \setlength{\abovecaptionskip}{-0.2mm}
    \includegraphics[width=2.8in]{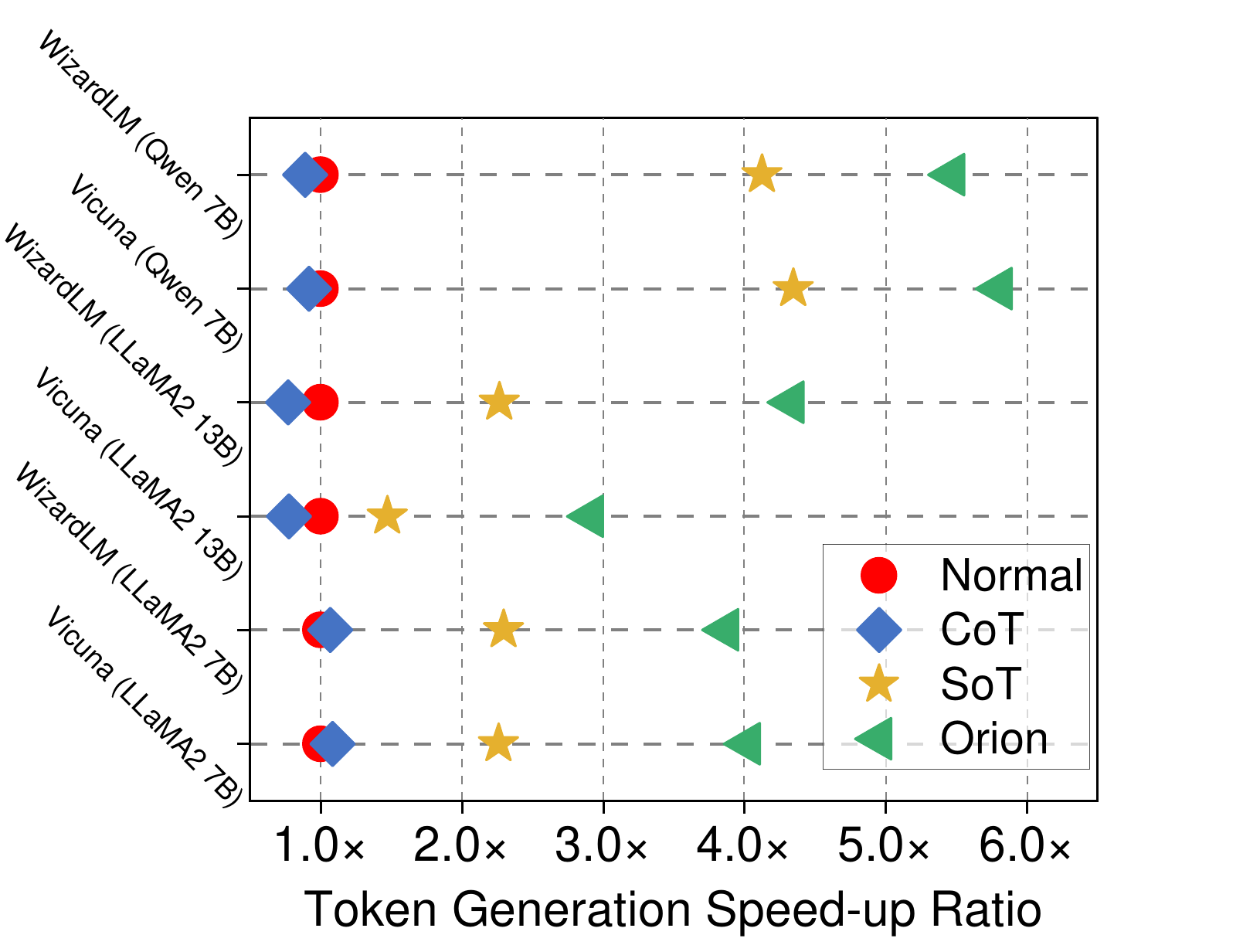}\label{subfig:init_chem_acc}
    \caption{Token generation speed-up ratio for Orion and baselines compare with Normal on the different datasets using various models.}
    \label{fig:init_experiment_chem}
    \vspace{-8mm}
\end{figure}

% \begin{figure}[t]
%     \centering
%     \setlength{\abovecaptionskip}{-0.2mm}
%     \subfigure[Test Accuracy]{
%         \includegraphics[width=1.6in]{fig/motivation/Speed_Up_Ratio.pdf}\label{subfig:init_chem_acc}
%     }\hspace{-0.3cm}
%     \subfigure[Communication Cost]{
%         \includegraphics[width=1.6in]{fig/motivation/Speed_Up_Ratio.pdf}\label{subfig:init_chem_communication}
%     }
%     %\setlength{\abovecaptionskip}{1pt}
%     \caption{Quality}
%     \label{fig:init_experiment_chem}
%     \vspace{-5mm}
% \end{figure}

\begin{figure}[t]
    \centering
    \setlength{\abovecaptionskip}{-0.2mm}
    \subfigure[GPT-5]{
        \includegraphics[width=1.6in]{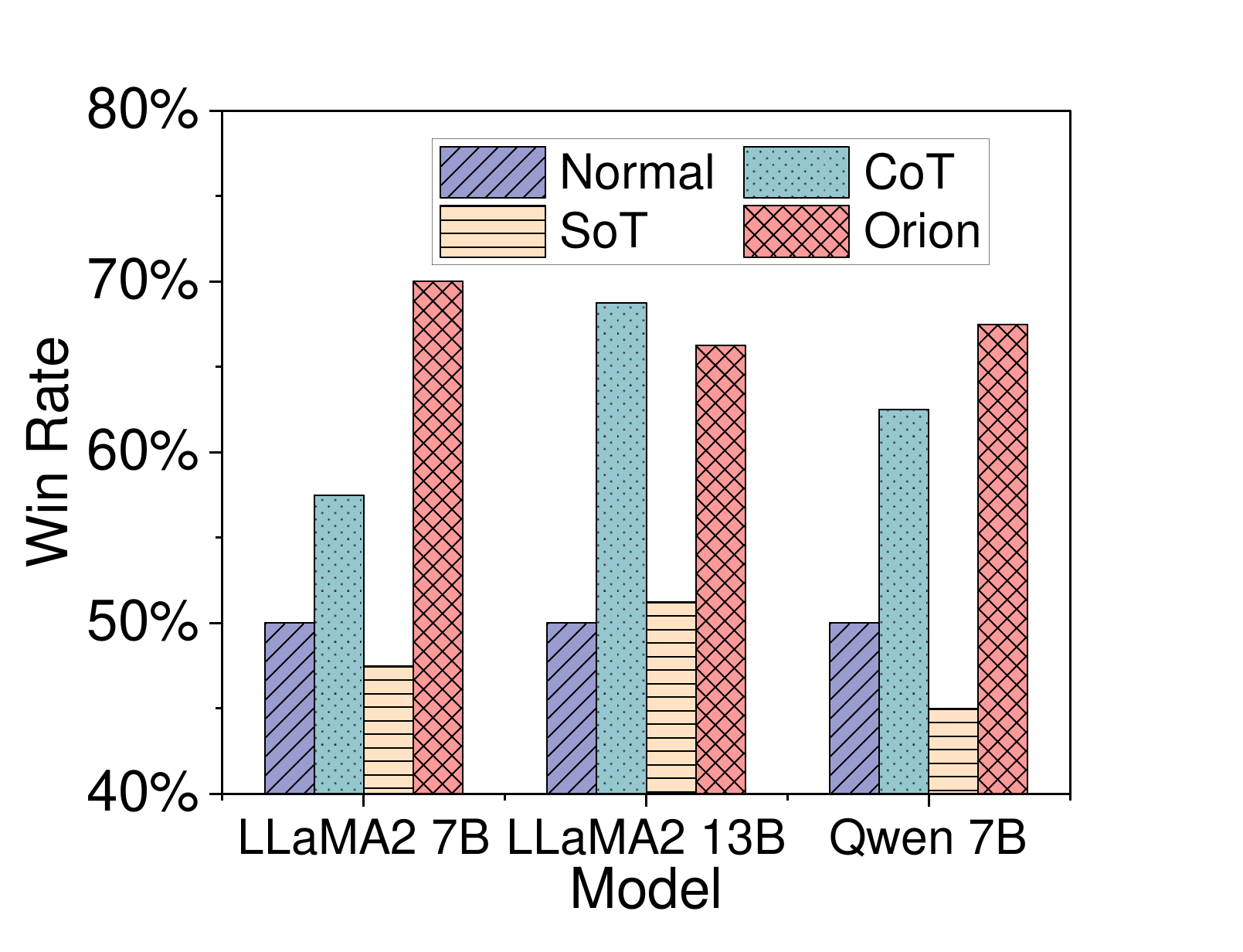}\label{subfig:init_sn_com}
    }\hspace{-0.3cm}
    \subfigure[DeepSeek]{
        \includegraphics[width=1.6in]{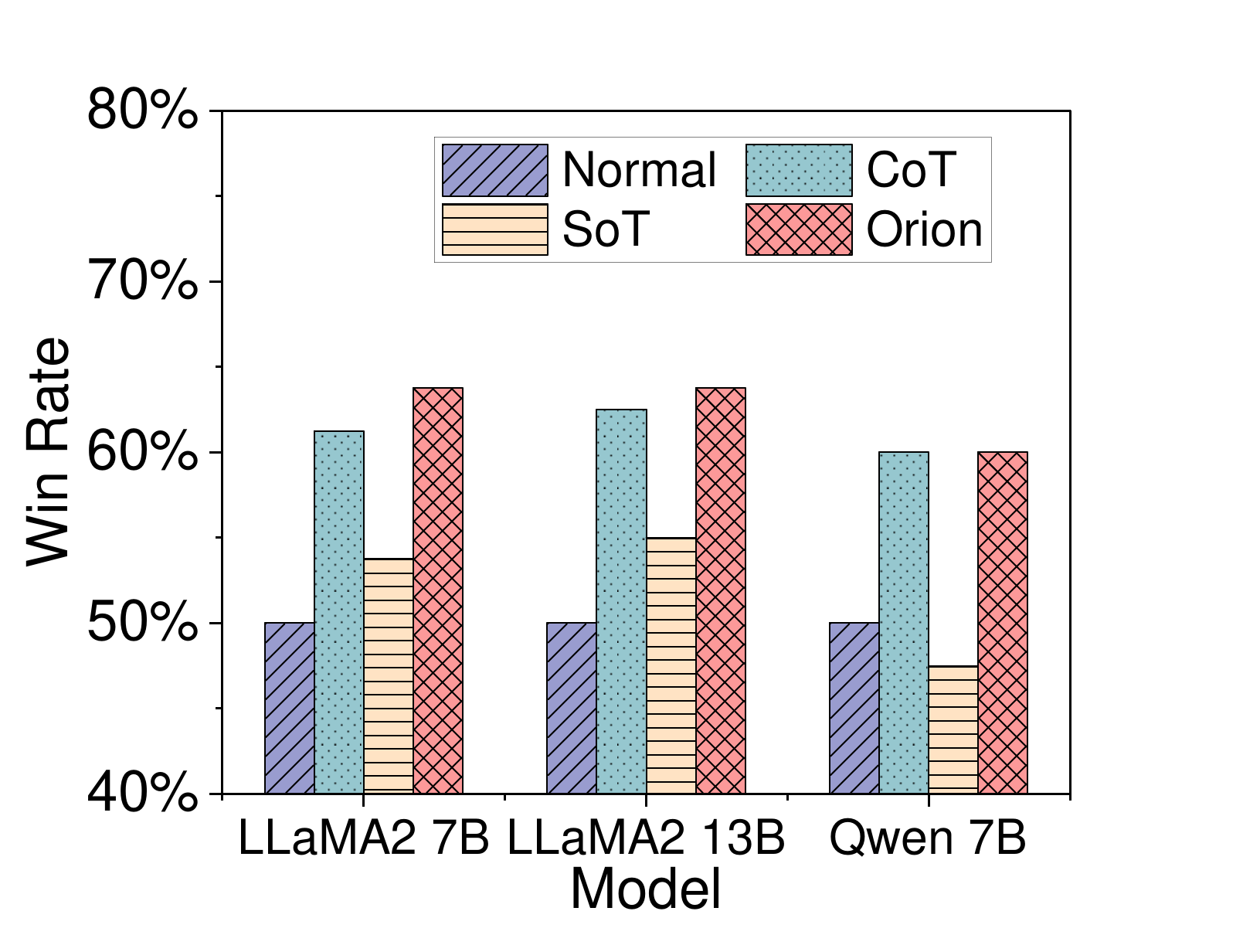}\label{subfig:init_sn_communication}
    }
    \caption{Quality for Orion and baselines with the two different judge models on the Vicuna dataset.}
    \label{fig:init_experiment_sn}
    \vspace{-7mm}
\end{figure}

% \begin{figure}[t]
%     \centering
%     \setlength{\abovecaptionskip}{-0.2mm}
%     \subfigure[Test Accuracy]{
%         \includegraphics[width=1.6in]{fig/motivation/Speed_Up_Ratio.pdf}\label{subfig:init_sn_com}
%     }\hspace{-0.3cm}
%     \subfigure[Communication Cost]{
%         \includegraphics[width=1.6in]{fig/motivation/Speed_Up_Ratio.pdf}\label{subfig:init_sn_communication}
%     }
%     %\setlength{\abovecaptionskip}{1pt}
%     \caption{Latency (zhuzhuang)}
%     \label{fig:init_experiment_sn}
%     \vspace{-5mm}
% \end{figure}

\subsection{Overall Performance Comparison}\label{subsec:Overall Comparison}
We begin by presenting the generation speed-up ratio of Orion and the baselines on two datasets, as shown in Figure  \ref{fig:init_experiment_chem}. The results demonstrate that Orion significantly outperforms all baseline methods. On the Vicuna dataset, Orion achieves a notably higher token generation speed-up ratio, surpassing Normal and CoT by 3.02$\times$ and 2.94$\times$ on the LLaMA2 7B model, respectively. 
Furthermore, Orion also exceeds the state-of-the-art method (\ie, SoT), increasing the average token generation speedup ratio from 2.69$\times$ to 4.25$\times$ across three different models. 
On the WizardLM dataset, since the queries cover a wider variety of types and exhibit greater difficulty variance, higher demands are placed on the generalization and adaptability of reasoning. 
Nevertheless, Orion still achieves an impressive 3.87$\times$ improvement in token generation speed-up ratio on the LLaMA2 7B model. Orion also outperforms SoT, which employs parallel optimization techniques, by 1.57$\times$. These findings indicate that Orion can substantially increase text generation speed across problems of varying categories and difficulty levels.
This consistent performance underscores Orion’s efficiency and adaptability in handling diverse and challenging query types. Moreover, the ability to maintain such speedup gains highlights the effectiveness of its query decomposition and parallel expansion.

\begin{table*}[htbp]
    \centering
    \caption{Latency for Orion and baselines on the WizardLM dataset.}
    \begin{tabular}{c|cc|cc|cc|cc}
        \hline
        \multirow{3}{*}{Method} & \multicolumn{4}{c|}{Vicuna} & \multicolumn{4}{c}{WizardLM}\\\cline{2-9}
        & \multicolumn{2}{c|}{LLaMA2 7B} & \multicolumn{2}{c|}{Qwen2.5 7B} & \multicolumn{2}{c|}{LLaMA2 7B} & \multicolumn{2}{c}{Qwen2.5 7B}\\\cline{2-9}
        & Time (S) & Speed-up Ratio & Time (S) & Speed-up Ratio & Time (S) & Speed-up Ratio & Time (S) & Speed-up Ratio \\\hline
        Normal & 2417 & 1.00$\times$ & 2837 & 1.00$\times$ & 5684 & 1.00$\times$ & 7829 & 1.00$\times$ \\\hline
        CoT & 2598 & 0.93$\times$ & 2900 & 0.98$\times$ & 6365 & 0.89$\times$ & 7957 & 0.98$\times$ \\\
        SoT & 1180 & 2.05$\times$ & 4921 & 0.58$\times$ & 4602 & 1.24$\times$ & 13694 & 0.57$\times$ \\\hline
        \textbf{Orion} & \textbf{742} & \textbf{3.26$\times$} & \textbf{843} & \textbf{3.37$\times$} & \textbf{2082} & \textbf{2.73$\times$} & \textbf{2292} & \textbf{3.42$\times$}\\\hline
    \end{tabular}
    \vspace{-8mm}
    \label{TABLE_latency}
\end{table*}

As shown in Figure \ref{fig:init_experiment_sn}, we present the test results on various models using the Vicuna dataset. It is noteworthy that all the methods are compared against Normal, with its win rate set at 50\% as the reference. Overall, Orion leads in quality on various models. For instance, with the LLaMA2 7B model, Orion achieves an overall win rate of 70\% in the answer quality using GPT-5 as the judge model, outperforming the CoT and SoT methods, which only reach 57.5\% and 47.5\%, respectively. Similarly, using Deepseek as the judge model, Orion still achieves a win rate of 63.75\%, while SoT only achieves 53.75\%. Furthermore, we have also observed that compared to DeepSeek, Orion performs better when ChatGPT is used as the judge model. This stems from ChatGPT's preference for rich and long outputs. However, even though the two judge models have their respective preferences, Orion still achieves optimal performance. These results underscore Orion’s strength in enhancing text generation quality compared to baseline methods. This significant improvement stems from our successful identification of dependencies between key points. By constructing a DAG to show the dependency relationship between key points, Orion guides the model to generate logically independent points in parallel while providing necessary contextual information for dependent points. This approach ensures both increased text generation speed and maintained text quality.

We evaluate the latency of Orion and the baseline at different models and datasets to present Orion's performance in Table \ref{TABLE_latency}. Overall, we find that Orion can significantly reduce the time required to complete all the queries. For instance, on the WizardLM dataset, Orion reduces latency by 2.42$\times$ compared to Normal when using the Qwen2.5 7B. It is noteworthy that we find SoT, although also employing a parallelization strategy, exhibits an increase in completion latency when using the Qwen2.5 7B model. 
This is because SoT cannot effectively constrain the content output, and the architectural design of the Qwen2.5 7B model also tends to output more tokens. Consequently, although SoT improves token generation speed (\eg, more than a 4.1$\times$ speed-up ratio), the longer output ultimately leads to an overall increase in completion latency. In contrast, Orion imposes strict constraints on the output, ensuring that fewer tokens can yield higher-quality outputs.

% \begin{figure}[t]
%     \centering
%     \setlength{\abovecaptionskip}{-0.2mm}
%     \subfigure[Test Accuracy]{
%         \includegraphics[width=1.6in]{fig/motivation/Speed_Up_Ratio.pdf}\label{subfig:init_sn_com}
%     }\hspace{-0.3cm}
%     \subfigure[Communication Cost]{
%         \includegraphics[width=1.6in]{fig/motivation/Speed_Up_Ratio.pdf}\label{subfig:init_sn_communication}
%     }
%     %\setlength{\abovecaptionskip}{1pt}
%     \caption{various (hengxiang)}
%     \label{fig:init_experiment_sn}
%     \vspace{-5mm}
% \end{figure}

\begin{figure}[t]
    \centering
    \setlength{\abovecaptionskip}{-0.2mm}
    \subfigure[Vicuna]{
        \includegraphics[width=1.6in]{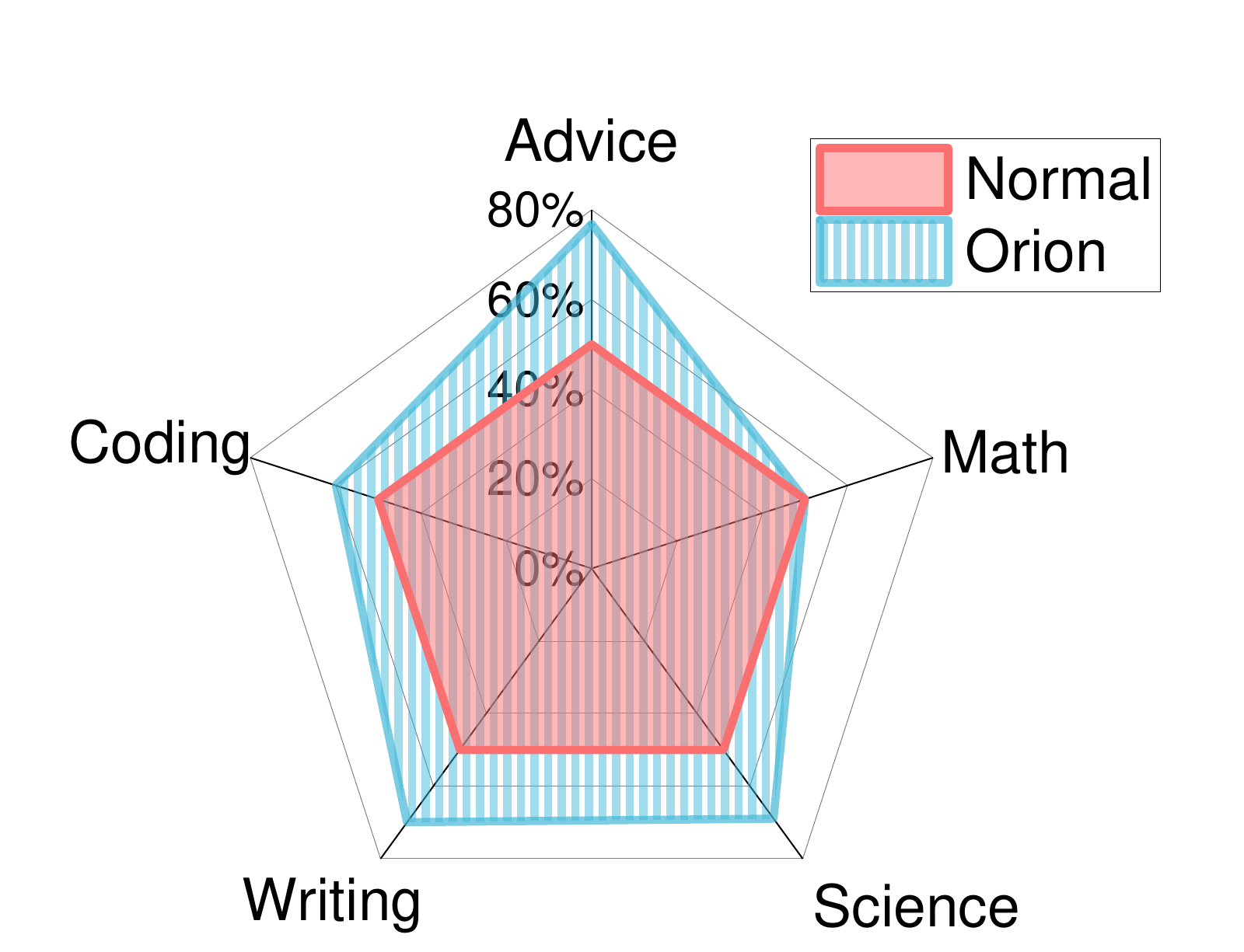}\label{subfig:init_sn_com}
    }\hspace{-0.3cm}
    \subfigure[WizardLM]{
        \includegraphics[width=1.6in]{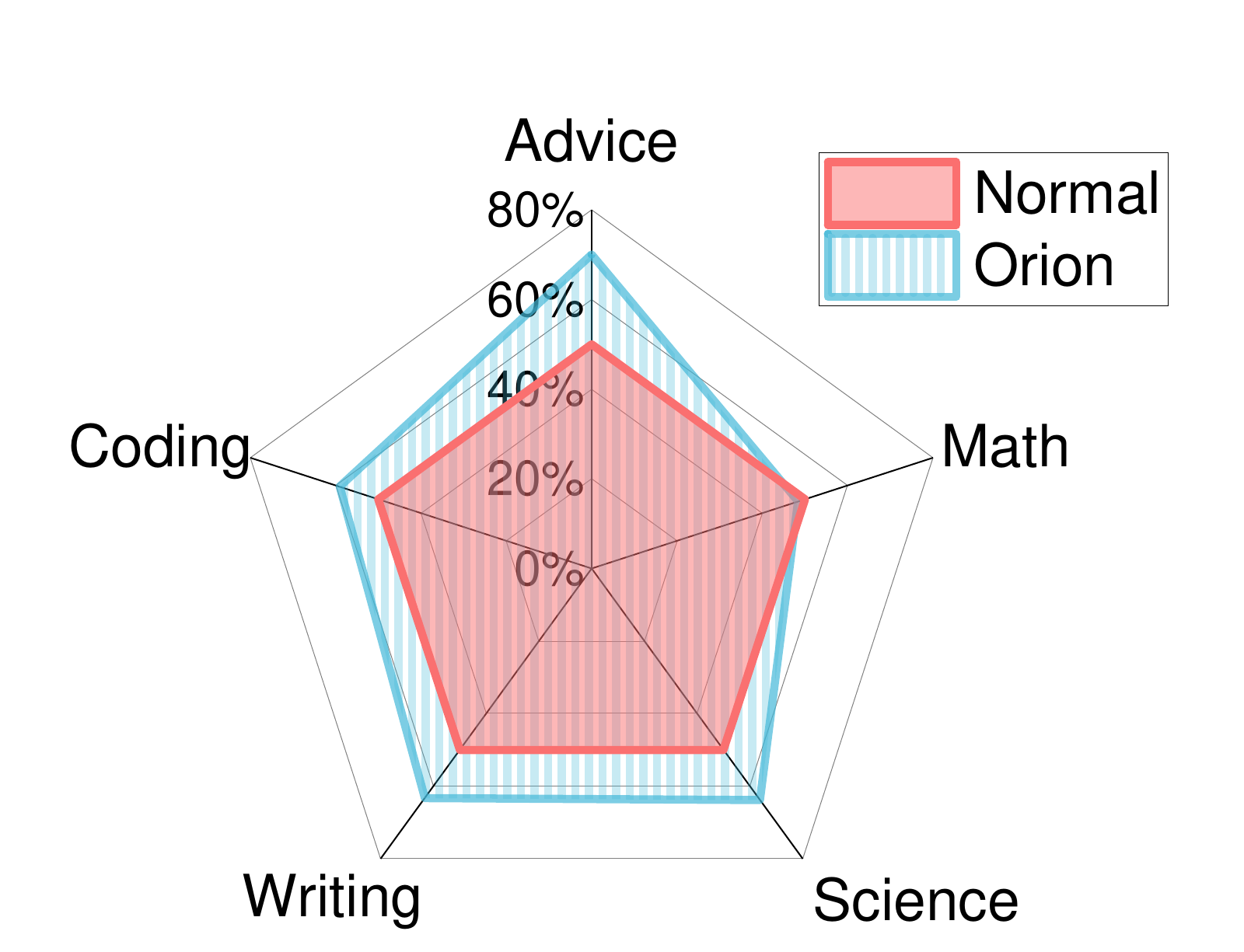}\label{subfig:init_sn_communication}
    }
    \caption{Quality for Orion and Normal on different categories queries with LLaMA2 7B.}
    \label{fig:categories_quality}
    \vspace{-8.5mm}
\end{figure}

\subsection{Performance Comparison with Different Query Categories}\label{subsec:Overall Comparison}
We evaluate the win rate of Orion and Normal (the balanced approach, CoT has disadvantages in efficiency, while SoT falls short in quality) across different question categories in various datasets, as shown in Figure \ref{fig:categories_quality}. Overall, Orion performs better in advice, coding, writing and science, while it slightly underperforms in math. Benefiting from its ability to identify key points in answering queries and determine dependency relationships between them, Orion excels particularly on queries that can be addressed from multiple perspectives, such as advice and science. Orion also maintains an advantage over Normal in queries requiring certain logical continuity, such as writing and coding. This advantage stems from the fact that, during writing queries, logical connections between key points exist but are not overly strong. When Orion encounters key points with dependencies during writing, it can reference the content of related points and continue writing, thus avoiding significant disruption. However, for queries demanding rigorous logical precision (\ie, math), even with outputs from other key points as references, Orion still has difficulty delivering excellent solutions and therefore slightly underperforms compared to Normal.

\begin{figure}[t]
    \centering
    \setlength{\abovecaptionskip}{-0.2mm}
    \subfigure[Vicuna]{
        \includegraphics[width=1.6in]{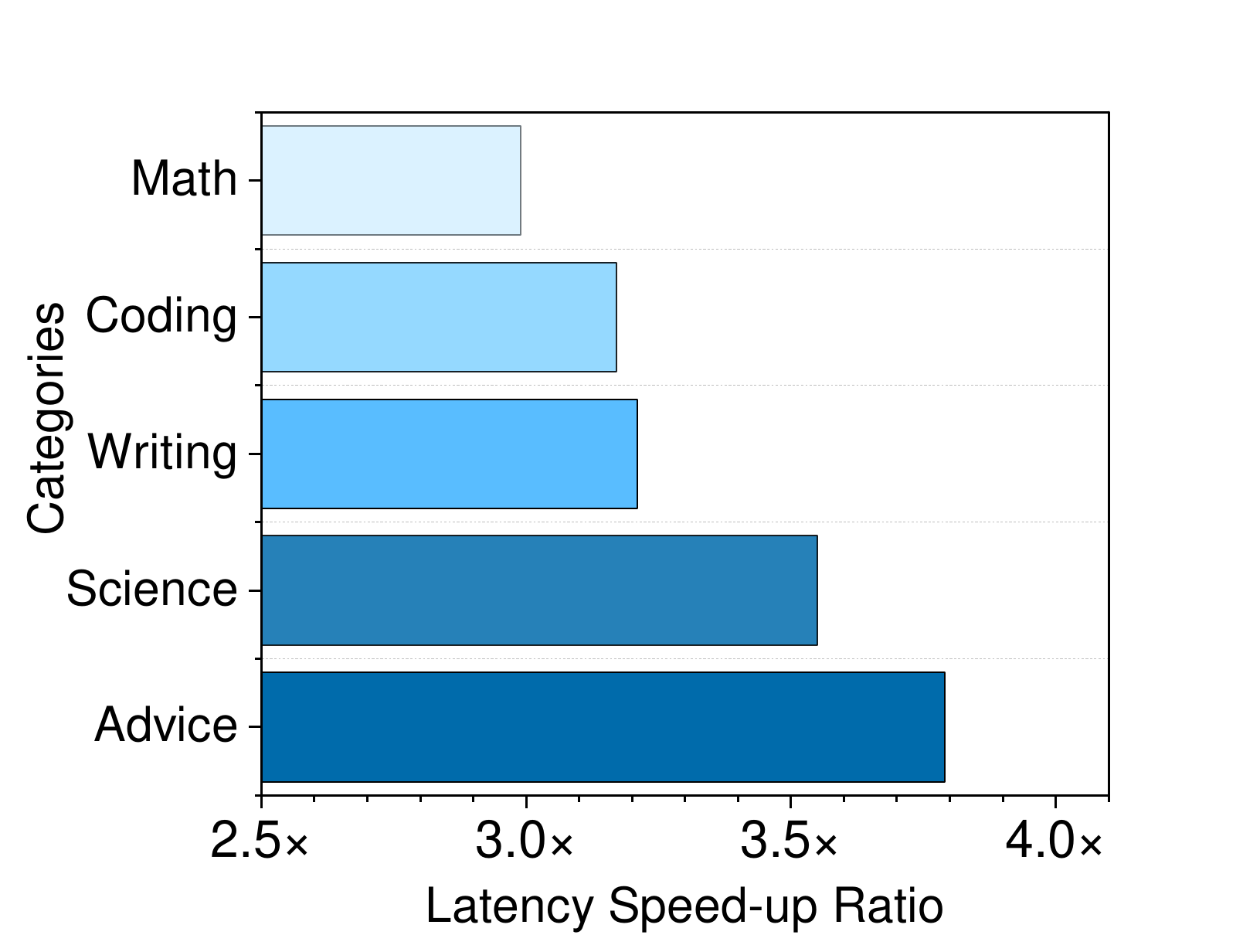}\label{subfig:init_sn_com}
    }\hspace{-0.3cm}
    \subfigure[WizardLM]{
        \includegraphics[width=1.6in]{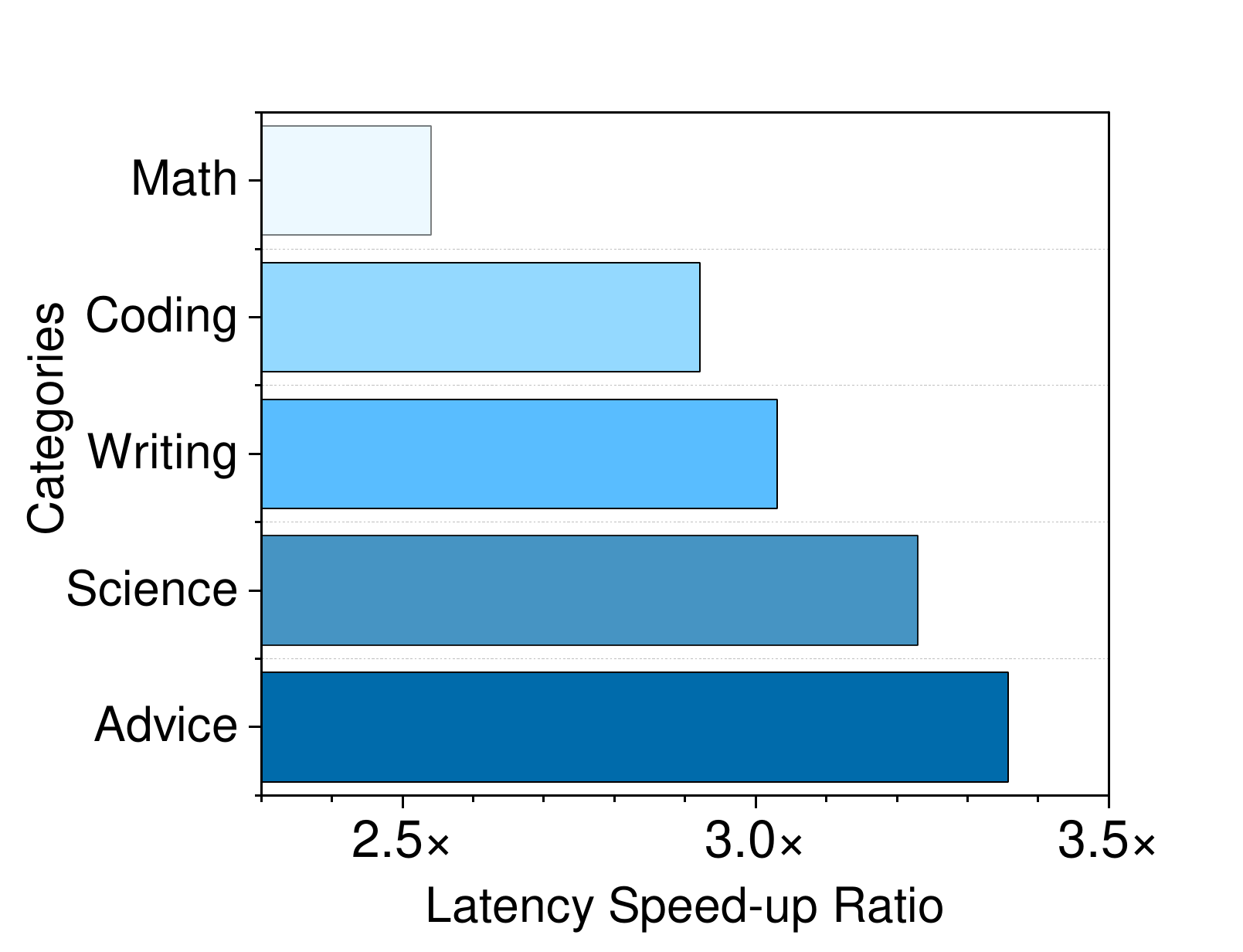}\label{subfig:init_sn_communication}
    }
    \caption{Latency speed-up ratio for Orion on different query categories using LLaMA2 7B compared to Normal.}
    \label{fig:categories_latency}
    \vspace{-8mm}
\end{figure}

Furthermore, we present the latency of Orion and Normal across different query categories in Figure \ref{fig:categories_latency}. Notably, Orion exhibits a stratified performance across various categories of queries. Orion achieves significantly better latency performance in categories such as advice and science compared to math. For example, on the Vicuna dataset, Orion improves advice queries latency by 3.79$\times$ compared to Normal, while for math-related queries, Orion improves by only 2.99$\times$. In categories like coding and writing, Orion's performance falls somewhere in between. Further analysis reveals that for advice and science queries, the generated key points are often independent or contextual, allowing for a significantly higher degree of parallelism during the subsequent content expansion phase. In contrast, for math-related queries, the key points are predominantly interdependent, meaning subsequent points require content decoded from earlier points to begin expansion. This dependency constrains the parallelism of content expansion, which in turn results in relatively higher latency.

% 我们首先展示了 Orion 及其基线在两个数据集上的文本生成加速率，如图 \ref{} 所示。结果表明，Orion  的表现显著优于所有基线。
% 在Vicuna数据集上，Orion  的文本生成加速率显著提升，与 XXX 和 XXX 相比分别提升了 XXX 和 XXX。此外，Orion 的表现也优于SOT，在三个不同模型上平均文本生成加速率提升了XXX。在wizardlm数据集上，由于数据集中问题包含了更多的种类且难度变化较大(从简单的速度计算到复杂的求解微分方程)，对于推理所需的泛化性和适应性提出了更加严格的要求。因此，大多数基线的性能显著下降。
% 然而，Orion 仍然实现了令人印象深刻的 XXX 的文本生成加速率提升。
% 它也比具有同样使用了并行优化技术的 SoT 高出 XXX。在表明Orion在面对不同类别不同难度的问题时都能大幅度提高文本生成速率。

% 在图 \ref{}} 中，我们展示了不同数据集在LLaMA2 7B模型上进行后的测试结果。值得注意的是，我们所有方法都是相比于Normal比较得到的胜率，我们设置Normal本身胜率为50%。总体而言，Orion在大部分数据集和模型上都保持能够保持质量优势。例如，在xxx数据集上，Orion 的总体问题回答质量取得了XXX的胜率，高于XXX和XXX方法的XXX，XXX。同样，在 xxx数据集上，虽然问题种类更多类别更加复杂，Orion 仍然也能够实现XXX的胜率而XXX只能实现XXX的胜率。此外，在使用XXX模型时，Orion取得了卓越的XXX胜率，而其他方法均无法达到XXX(稍低一些的整数)胜率。这些结果凸显了 Orion在提升文本生成质量上的优势，相比基线方法更是如此。这一显著的提升来源于我们成功确定了生成要点间的依赖关系，Orion构建DAG图引导模型并行生成逻辑上独立的要点并且为存在依赖关系的要点提供相应的信息，由此在保证生成文本速率的同时维持了文本生成质量。

% 我们评估 orion 和normal在不同评价指标下的表现，以展示orion 可以在哪些方面改进或者损害了答案的质量。评价指标包括：连贯性，相关性，创新性，丰富性以及沉浸性。平均而言，我们发现Orion提升了创新性，丰富性以及相关性，但是却略微损害了沉浸感和连贯性。由于Orion采用多个要点来共同组成答案，因此在要点生成时，模型可以从不同的角度生成更加丰富多样的要点来回答问题，同时在扩展时对于每个要点都是独立进行扩展，因此在回答创意以及丰富性上有了更好的表现。然而，虽然在扩展时将具有依赖关系的要点的信息作为补充，但是这种方式难以保证模型对于并行生成点的连贯性，因此在连贯性和沉浸性上略微逊色。

% 进一步的，我们评估了Orion 和 Normal在所有问题类别中的净胜率表现在Fig. \ref{}中。与Fig. \ref{}类似，我们发现Orion在XXX上评估的表现比XXX更加乐观一些。整体上来看，Orion在XXX,XXX,XXX,XXX,XXX上表现更加出色一些而在XXX，XXX，xxx,XXX上略微逊色于Orion，在xxx,xxx,xxx上则表现类似。受益于Orion能够确定问题回答的要点并且确定要点间的依赖关系，Orion在能够从多个角度回答的问题上表现更加出色，例如角色扮演，通用等问题上。在具有一定逻辑连续性的问题上，Orion也能保持一定的优势相比于Normal，例如在写作问题上。这种优势来源于在生成写作要点时，每个要点间的逻辑行存在但不够强烈。Orion在执行写作任务时当遇到存在依赖关系的要点时，可以将对应要点的要点内容作为参考继续书写，因此并不会受到太大的影响。但是在面对诸如数学类逻辑严谨的问题时，即便依赖其他要点的输出结果，Orion仍然难以完全正确的给出问题的解，因此略逊于Normal。

\section{CONCLUSION}\label{sec:conclusion}
% In this work, we propose FedGCF to adaptively select clients to address the challenges of limited resources and the non-IID problem raised by FGL, for efficient recommendation systems. It employs an MAB-based online learning algorithm to adaptively determine the number of participating clients and select the clients that meet the criteria in each round. The dynamic optimization of client selection can contribute to the balance between the global model training performance and resource efficiency. The extensive experimental results demonstrate that FedACS significantly outperforms the existing baselines.

In this work, we propose Orion, a novel reasoning framework for LLM that decomposes complex queries into two synergistic phases: key point generation and content parallel expansion, significantly improving token generation speed and reducing latency. Orion introduces a dependency-aware parallel expansion algorithm that models inter-point relationships using a DAG, enabling parallelized content expansion while maintaining logical coherence and structural integrity in the generated answers. This approach effectively achieves both efficiency and reasoning quality. Extensive experiments demonstrate that Orion outperforms existing baselines in both efficiency and quality, achieving notable gains in token generation speed while reducing reasoning latency.

\bibliographystyle{IEEEtran}
\bibliography{papercontent/refs}

\end{document}